%% file: main.tex
\documentclass[10pt,twocolumn,letterpaper]{article}

\usepackage[pagenumbers]{cvpr} 

\input{preamble}

\definecolor{cvprblue}{rgb}{0.21,0.49,0.74}
\usepackage[pagebackref,breaklinks,colorlinks,citecolor=cvprblue]{hyperref}
\usepackage[capitalize]{cleveref}

\def\paperID{8699} 
\def\confName{CVPR}
\def\confYear{2024}

\title{Segmentation-Based Parametric Painting}

\author{Manuel Ladron de Guevara\\
Carnegie Mellon University\\
{\tt\small rldg.manuel@gmail.com}
\and
Matt Fisher\\
Adobe Research\\
{\tt\small matfishe@adobe.com}
\and
Aaron Hertzmann\\
Adobe Research\\
{\tt\small hertzman@dgp.toronto.edu}
}

{%
\begin{figure*}[!t]%
}{\end{figure*}}

\begin{document}

\input{figures/fig_tex/teaser}

\input{sec/0_abstract}    
\input{sec/1_intro}
\input{sec/2_related_work_v2}
\input{sec/3_method}

\input{sec/4_experiments}
\input{sec/5_conclusion}

{
    \small
    \bibliographystyle{ieeenat_fullname}
    \bibliography{main}
}


\end{document}

%% file: preamble.tex
\usepackage[dvipsnames]{xcolor}
\usepackage{times}
\usepackage{epsfig}
\usepackage{graphicx}
\usepackage{amsmath}
\usepackage{amssymb}
\usepackage{booktabs}

\usepackage{multirow}
\usepackage{float}
\usepackage{placeins}
\usepackage{adjustbox}
\usepackage{capt-of}
\usepackage{url}
\usepackage{bbding}
\usepackage{pifont}
\usepackage{caption}
\usepackage{subcaption}
\usepackage{tabularx}
\usepackage{array}

\usepackage{listings}
\usepackage{algorithm}
\usepackage{algpseudocode}

\newcommand{\loss}[1]{\mathcal{L}_{\mathrm{#1}}}

%% file: figures/fig_tex/teaser.tex
\twocolumn[{%
\renewcommand\twocolumn[1][]{#1}%
\maketitle
\begin{center}
    \includegraphics[width=\textwidth]{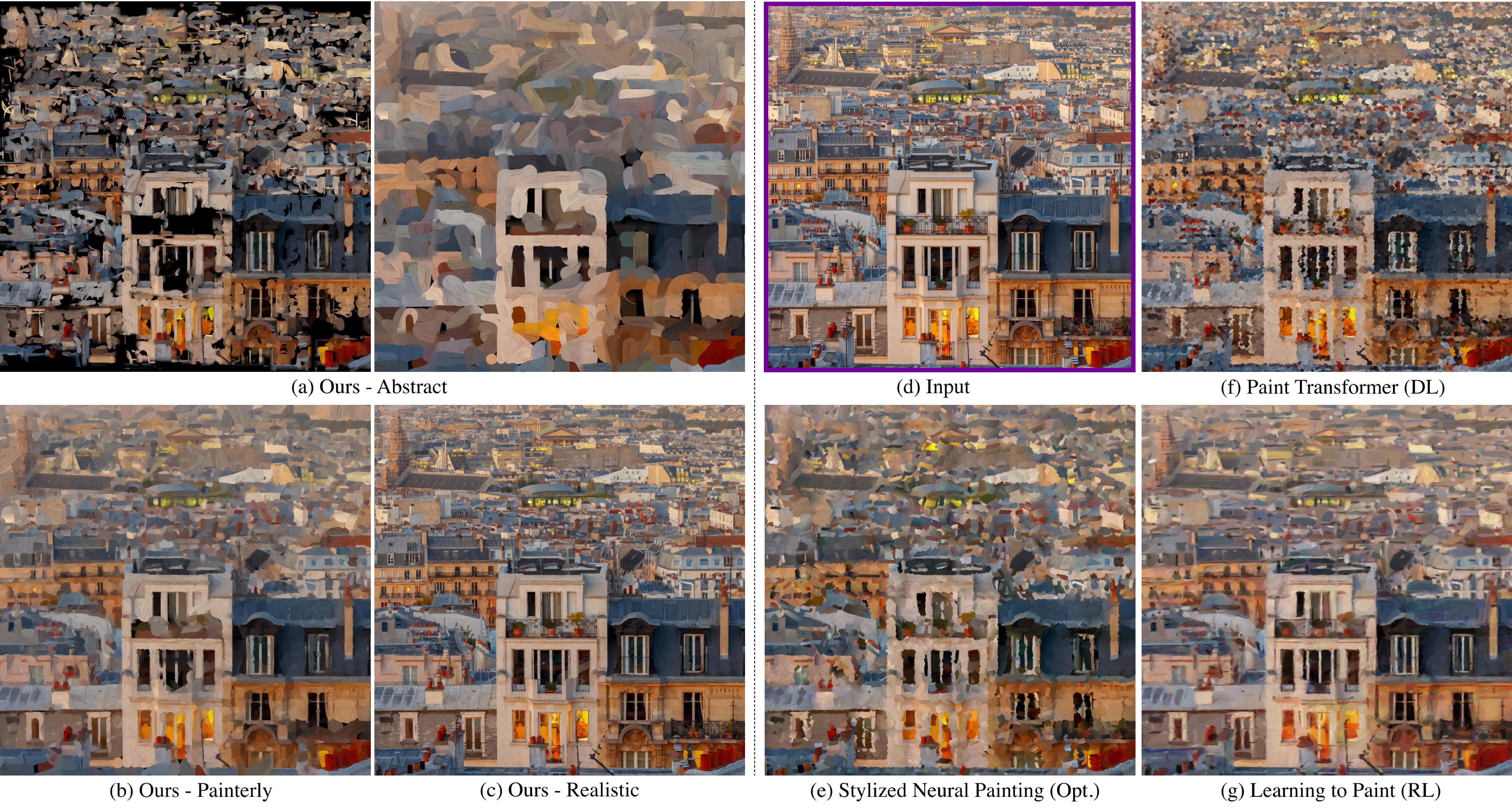}
    \captionof{figure}{
Comparison of our method against previous work on parametric painting algorithms. Panels (a) through (c) demonstrate our algorithm's capability to render scenes in various styles: abstract with simplified forms and abstract with a heavy emphasis on thick loose strokes, texture, and a lack of clear form (a), painterly impressionistic style with dynamic strokes and emphasis on effect (b), and realistic style capturing both micro and macro structures with high fidelity (c). These are juxtaposed against existing methods: Stylized Neural Painting (SNP) \cite{Zou_2021_CVPR} (e), Paint Transformer (PT) \cite{Liu_2021_ICCV} (f), and Learning to Paint (L2P) \cite{huang2019learning} (g), which exhibit limitations such as imprecise edge depiction and a lack of stylistic diversity. SNP and PT use rigid, blocky strokes that are unable to capture fine details, edges and shapes. L2P captures details better, although yields blurry results with visible seams at large resolutions. The input image is displayed in panel (d) for reference. This comparison underscores our method's superior ability to balance detail retention with stylistic expression. Zoom-in is encouraged to see the nuances of each work.}
    \label{fig:teaser}
\end{center}
}]

%% file: sec/0_abstract.tex
\begin{abstract}
\vspace{-1\baselineskip}
We introduce a novel image-to-painting method that facilitates the creation of large-scale, high-fidelity paintings with human-like quality and stylistic variation.
To process large images and gain control over the painting process, we introduce a segmentation-based painting process and a dynamic attention map approach inspired by human painting strategies, allowing optimization of brush strokes to proceed in batches over different image regions, thereby capturing both large-scale structure and fine details, while also allowing stylistic control over detail.
Our optimized batch processing and patch-based loss framework enable efficient handling of large canvases, ensuring our painted outputs are both aesthetically compelling and functionally superior as compared to previous methods, as confirmed by rigorous evaluations. 
Code available at: \href{https://github.com/manuelladron/semantic\textunderscore based\textunderscore painting.git}{https://github.com/manuelladron/semantic\_based\_painting.git}

\end{abstract}
\vspace{-1\baselineskip}


%% file: sec/1_intro.tex
\section{Introduction}
\label{sec:intro}

Painting is the world's oldest technology for creating visual imagery, and a continuing source of fascination and artistic expression. 
Representational painting is a complex process where an artist creates a picture of a realistic scene from a series of brush strokes. Artists use a variety of strategies to determine where and how to paint, and different strategies produce different painting styles. For example, a painter might work in a coarse-to-fine manner, beginning with general shapes and then refining details, or in a semantic-region-based manner, painting areas such as sky, mountains, and buildings sequentially; the latter strategy may produce a more segmented style than the former, with different styles for different objects.

These strategies, in part, are a function of human working memory and foveal vision: unlike optimizers, artists cannot simultaneously work on every part of a picture in parallel. They work piece-by-piece, focusing on one region or aspect of a painting at a time \cite{cohen2005look,glazek2012visual,perdreau2015drawing,tchalenko2009eye, tchalenko2009segmentation, chamberlain2016genesis}, before moving on to the next.


The computational \textit{stroke-based painting} problem is: given an input photograph, generate a painting to represent the photograph, parameterized by a set of brush strokes, which themselves are parameterized by their colors, control curves and thicknesses, with optional texturing \cite{Haeberli,hertzmannSBR}. These methods allow styles to be defined directly in terms of brush stroke properties, with further additional user control possible on the strokes themselves, e.g., \cite{ODonovanAniPaint}. 
Current methods treat this as an optimization problem: using a limited set of brush strokes, approximate the input image as closely as possible \cite{ganin2018synthesizing,Hertzmann:PBR,mellor2019unsupervised,huang2019learning,Zou_2021_CVPR,Liu_2021_ICCV,frida}. By varying the level of approximation, a user should be able to select from more- or less- abstract styles, and selecting different placement strategies should produce qualitatively different styles.  Moreover, robotic painting systems require explicitly-defined brush strokes, e.g., \cite{frida,edavid,arts7040084,bidgoli_ladron}, which are not available with pixel-based stylization techniques.



Despite recent progress, existing methods for this problem suffer from a limited ability to perform this optimization well for large images and limited user control. State-of-the-art methods either jointly optimize a large collection of strokes, or train a network to generate them. These approaches are limited by the problem of generating strokes uniformly for an input image, which becomes impractical as the input size increases. Hence, these methods do not generate high-fidelity results for even moderate-sized input images, thereby limiting the level of abstraction and control \cite{huang2019learning, Liu_2021_ICCV, Zou_2021_CVPR, singh2021combining}.
Moreover, they lack mechanisms to mimic varied artistic strategies, such as applying  different styles for different scene elements, and coarse-to-fine painting. 

Inspired by these observations, we propose a new approach to generating stroke-based paintings from photographs.
We first structure the entire painting process in layers by isolating different semantic parts of the canvas. Then dynamic attention maps guide the painting process to refine areas that need more detail, and differentiable strokes are optimized sequentially within each region.   This sequential optimization process allows us to generate strokes over a large image canvas efficiently, in contrast to methods that produce pictures as a single joint optimization.
It also approximates human painting strategies that operate sequentially on different image regions, rather than jointly optimizing all strokes at once. 
Several objective function parameters are provided to control the style, including limiting the number of strokes and level of detail for each semantic region, providing a continuum from realism to abstraction.
Quantitative and qualitative evaluations show that our method achieves a greater range of high-fidelity reconstructions to abstract imagery, while also producing high-quality results at large image sizes.


    

    


%% file: sec/2_related_work_v2.tex
\input{figures/fig_tex/framework}
\section{Related Work}
\label{sec:related_work}

Early stroke-based-rendering (SBR) approaches varied from procedural, rule-based systems without optimization \cite{Haeberli,Litwinowicz}, including layering strokes in a coarse-to-fine fashion. 
Procedural methods effectively produce captivating paintings, but their rigidity in stroke placement is a notable limitation. In principle, optimization-based SBR overcomes this constraint, offering greater adaptability in stroke positioning and more accurately encoding artist choices \cite{Hertzmann:PBR}. Such methods have evolved from heuristic-based optimization to more advanced techniques such as EM-like packing algorithms for arranging non-overlapping stroke primitives \cite{hertzmannSBR,NPRbook}, with applications extending to 3D model stylization \cite{meier1996painterly,overcoat}. Recent SBR methods employ deep networks trained to simulate a painting agent through reconstruction losses, involving pixel and perceptual losses \cite{johnson2016perceptual,zhang2018perceptual}, facilitated by differentiable rendering engines \cite{li2020differentiable}. However, the difficulty of optimizing large collections of strokes over an entire image remains challenging.

Reinforcement Learning (RL) has become prominent in SBR for simulating decision-making in painting agents \cite{xie2013rl,ganin2018synthesizing, mellor2019unsupervised, huang2019learning, lpaintb, singh2021combining,schaldenbrand}, with diverse applications ranging from detailed reconstructions to robotic painting. SPIRAL \cite{ganin2018synthesizing} and its enhanced version SPIRAL++ \cite{mellor2019unsupervised} are noteworthy for their adversarial training algorithms, despite producing less defined images. Nonetheless, both methods struggle with interpretability, style control, and accurate input reconstructions. Some RL methods focus on accurate depiction \cite{huang2019learning, singh2021combining}. Huang \textit{et al.}'s method
\cite{huang2019learning} needs thousands of tiny strokes to reconstruct an image, although becomes blurry for high resolution images, and it is stylistically limited. 

Deep learning methods like Paint Transformer \cite{Liu_2021_ICCV} utilize a CNN-Transformer architecture to predict stroke sequences, showing promise in generalization despite pattern repetition in results. RNN-based models have also been explored for distinct line drawing and sketching \cite{ha2017neural,kingma2013auto, zheng2018strokenet, mo2021virtualsketching}, with our method adopting similar attention mechanisms \cite{mo2021virtualsketching}.

Additionally, recent works have leveraged direct differentiable optimization for stroke placement \cite{li2020differentiable,Zou_2021_CVPR}, with style variations often induced by style-transfer techniques in pixel space \cite{Gatys_2016_CVPR, johnson2016perceptual}.



%% file: figures/fig_tex/framework.tex
\begin{figure*}[t]
\begin{center}
\includegraphics[width=\textwidth]{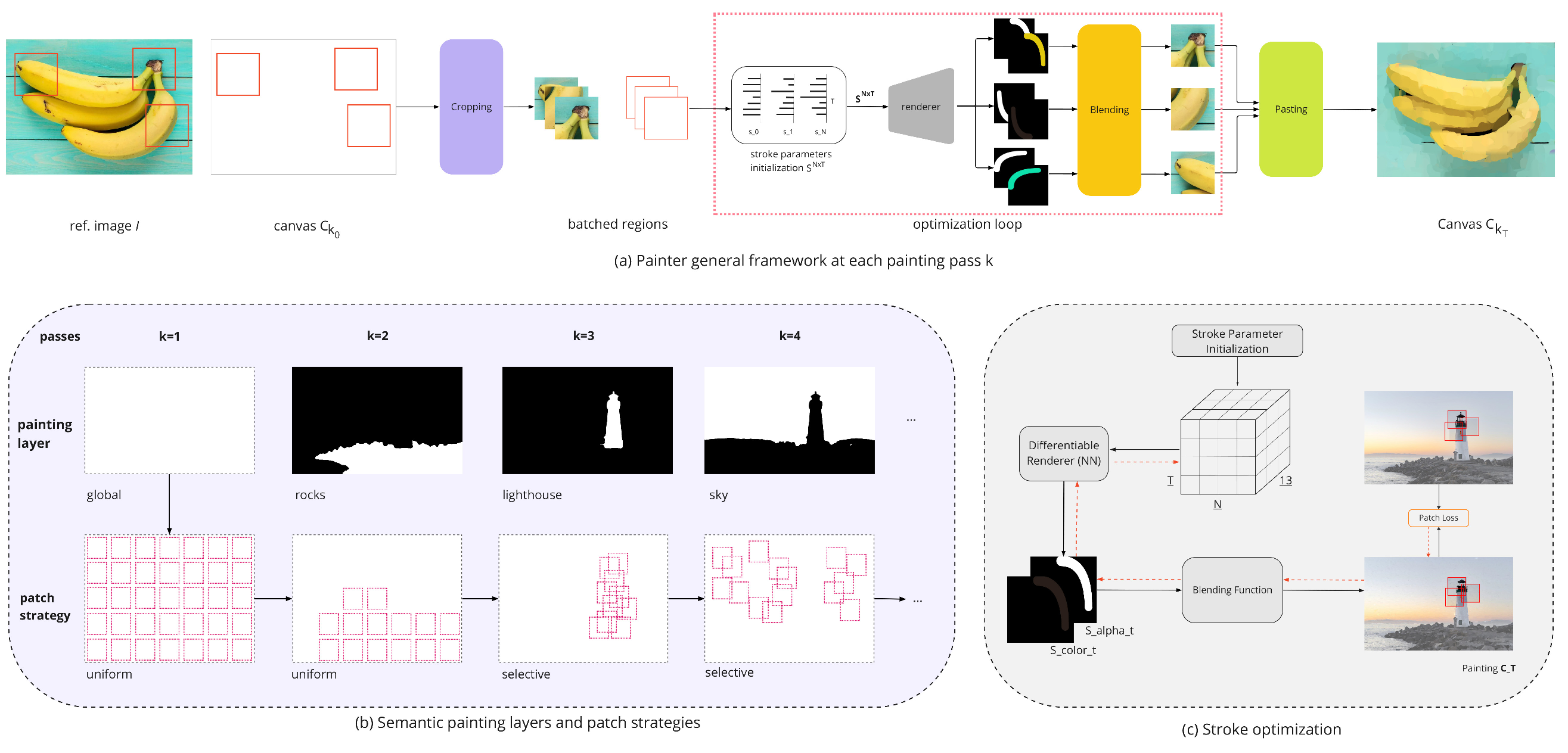}
\end{center}
\vspace{-1em}
\caption{Overall Painting Framework. Our framework models the painting process as a set of sequential painting passes. (a) Pipeline of each painting pass $k$, which is composed by four parts: attention patches, cropping, optimization, and pasting. (b) Semantic layers and patch strategies. (c) Stroke optimization module with patch-loss.}
\label{fig:framework}
\vspace{-1em}
\end{figure*}

%% file: sec/3_method.tex
\section{Method}
\label{sec:method}

Our goal is to efficiently optimize a collection of stroke parameters that define a large-scale painting from a given image, capturing micro and macro structures with high fidelity, and providing control  for stylistic variation. Our system takes as input an image $X$, and style parameters, encapsulated by three style presets: ``realistic,'' ``painterly,'' and ``abstraction.'' Then,  our objective is to optimize a parameter vector $\bar{Y}$ of strokes, which, when rendered on a canvas $C$, will approximate the given image $X$ as a stylized painting. Our scalable solution accommodates varying sizes and aspect ratios through batch-patch optimization, cropping the input and canvas into a set of $128\times128$ patches, which are reassembled after the optimization process to match the original image scale. 




Inspired by human painting techniques, and to provide more control over the painting process, we first use a segmentation network that decomposes the painting into semantic areas, and then employ dynamic attention maps to determine where to paint. An overview of our framework is shown in \Cref{fig:framework}. For information on the parametrization of styles and implementation details, refer to supplemental.

\subsection{Layered Painting and Patch-Based Optimization}

The set of paint strokes is separated into $K$ layers, one for each input semantic region. Applying separate painting styles or details for different regions allows stylistic control over separate scene elements. 
For each painting layer, the canvas is divided into a set of $128\times128$ patches, where all stroke parameters in each patch are batch-optimized; different layers may have different numbers of patches, as a function of the semantic layer's area. 


The optimization is performed in a series of $P$ passes, designed to  progress from general, broader strokes to intricate, finer details, akin to the approach of a traditional artist \cite{Hertzmann1998}. For each painting pass $p$, our method applies an optimization phase to each of the $K$ semantic layers, following a coarse-to-fine procedure. 

Let $P$ denote the total number of painting passes, and let's denote the number of semantic layers as $K$. We define $k_p$ as the layer index being refined in pass $p$, where $k_p \in \{0,1,2,...,K-1\}$. 
Overall, our goal is to minimize a reconstruction loss $L_{rec}$ summed over the entire image. We specify this loss later in the paper.
At a given pass, the goal is to optimize just the strokes that appear in some layer $k$. The objective for each layer $k$ during optimization is to optimize a subset of the loss just for the patches and strokes in this layer: $\arg \min_{\bar{y}_{k}} \sum_{i=1}^{N_{k}} \mathcal{L}_{rec}(X_{k,i}, C_{k,i}(\bar{y}_{k}))$, where $X_{k,i}$ is the $i$-th patch of the semantic region $k$ of the input image, and $C_{k,i}(\bar{y}_{k})$ denotes the $i$-th patch of the canvas at layer $k$, parameterized by the stroke parameters $\bar{y}_{k}$. The number of layers $K$ adapts to each painting based on the unique semantic content of the input image, and it is determined automatically by the segmentation network. In cases where the input image lacks semantic distinction, or if semantic assistance is disabled, a single layer approach is adopted ($K=1$), involving the entire canvas. We use the DETR model \cite{carion2020end}, which consists of a CNN (ResNet) backbone followed by an encoder-decoder Transformer, trained on the COCO Panoptic Segmentation task.  



Starting with the input image $X$, we derive all segmentation layers and compute each binary mask, which is then used to initialize the vector of stroke parameters $\bar{Y}_{k_p}$. 
Upon the completion of the optimization steps, strokes that fall outside the designated segmented area are filtered out based on a threshold $\gamma$. 

\paragraph{Dynamic Attention Maps.}
While painting, any human painter must continually decide where next to paint brush strokes, and different painters will make these choices differently.
Our system represents this process by dynamic attention maps, with  
 two distinct strategies: uniform, and selective (see \Cref{fig:framework}(b)). The former is based on a uniform distribution across the canvas, yielding patches in a consistently-spaced grid covering the entire image. 
The latter identifies the $V$ patches with the highest error for attention, where $V$ is a user-determined style parameter. This error is computed as $L_1$ loss between the canvas and image patches. We found that just using $L_1$ loss is enough to provide a good painting guidance, making the process more efficient in comparison to using perceptual losses. 


The uniform approach ensures an unbiased and uniform distribution of detail throughout the painting, and is typically reserved for the foundational stages of a painting, generally producing higher realistic painting than the selective approach, which generates a more human-like, organic painterly style paintings. 


\paragraph{Stroke Initialization, Renderer, and Blending.}
We use the stroke parameter representation and differentiable renderer, $G$, provided by \cite{huang2019learning} (for more details refer to supplemental).
Each stroke is parameterized by a 13-dimensional tuple $ \{x_0,y_0,x_1,y_1,x_2,y_2,r_0,r_2,t_0,t_2,R,G,B\}$ that encodes the start, middle and end points of a quadratic Bézier curve, radii and transparency at the start and end points, and RGB color of the stroke. For each $128 \times 128$ patch in the canvas, we distribute $T$ stroke parameters evenly in a regular array. 


Strokes are composited into a patch with soft blending: $M \leftarrow M_{t-1} \odot (1-G(s_j)_a) + G(s_j)_c$ where $s_{j}$ are the parameters of the $j$-th stroke, $M$ is a canvas patch, and $G(s_j)_a$ and $G(s_j)_c$ are the alpha channel and the stroke given by the differentiable renderer $G$, respectively. However, for semantic layers, 
 the corresponding binary mask $\alpha_k$ is used: $M_k \leftarrow  M_{k_{t-1}} \odot (1 - G(s_j)_a) + G(s_j)_c \odot \alpha_k $


\subsection{Optimization and Loss Functions.}
At a given pass $p$, for each layer $k$ in such a pass, we optimize all stroke parameters from all $N_{k}$ patches in batch. That is, we optimize a matrix $S_{k}^{N\times T \times 13}$ altogether, where $T$ is the stroke budget, and instead of computing the loss function over the entire canvas, we use a patch loss described earlier (see \Cref{fig:framework} (c)). This allows us to perform these optimizations much more efficiently than optimizing over the entire canvas. Formally, given a set of canvas patches and the same set of reference image's patches, our overall loss function is:

\begin{equation}
\label{eq:loss}
    \mathcal{L}_{rec} = \frac{1}{N} \sum_{n=0}^N \alpha \loss{pixel_n} + \beta \loss{perc_n}
\end{equation}
where $N$ is the total number of patches. For our pixel loss, we use the $L_1$ function, $\loss{pixel} = |C - I|$, where C is a canvas and I a reference image. For perceptual loss, we use the feature vectors from a pretrained VGG16 \cite{zhang2018perceptual}. Specifically, let $V_{ij} = \{V_{ij}^1, ..., V_{ij}^k\}$ and $W_{ij} = \{W_{ij}^1, ..., W_{ij}^k\}$ be a set of $k$ feature vectors extracted from image $I$ and canvas $C_T$, respectively. We use cosine similarity as follows:

\begin{equation}
\label{eq:perc}
     \loss{perc} = \cos{\theta} = \sum_k^K \sum_{i,j} \frac{V_i W_j}{\left\|V_i\right\| \left\|W_j\right\|}
\end{equation}

where ${i,j}$ index the spatial dimensions of the feature maps $V$ and $W$, and $K$ the extracted layers from VGG16 trained on ImageNet \cite{simonyan2014very}. We use layers 1, 3, 6, 8, 11, 13, 15, 22 and 29, and optimize via gradient descent and use Adam optimizer \cite{kingma2014adam} to find the set of strokes $S_{k_p}$.

%% file: sec/4_experiments.tex
\input{figures/fig_tex/coarse2fine}
\vspace{-1em}
\section{Experiments}
\label{sec:exp}

We now demonstrate stylistic effects achievable with our method, and compare with state-of-the-art methods.

The semantic control provided by our method allows a user to apply different styles to different regions, akin to classical artistic techniques where backgrounds might be abstracted for emphasis on detailed foregrounds and vice-versa, reminiscent of techniques observed in classical artworks \cite{techniques}, and specific subjects or objects may be highlighted. 
The versatility of this approach, resulting in varied visual interpretations based on the areas of emphasis, is illustrated in \Cref{fig:teaser} (a) and (c), \Cref{fig:biases}, \Cref{fig:painterly_vs_realistic},  \Cref{fig:control}, and \Cref{fig:sota_comp_2}.



\input{figures/fig_tex/biases}
\subsection{Style Variations}
Painting techniques and strategies lead to different styles, and our algorithm can output a set of stylistic outputs, ranging from realism to abstraction and painterly styles.

\input{figures/fig_tex/painterly_vs_realistic}

\paragraph{Realism.}
\input{figures/fig_tex/control}

We begin by demonstrating a style that precisely reproduces the target image.  Previous methods cannot achieve accurate image reproduction through optimization; this limits their ability to define new styles, since many styles cannot be accurately optimized, e.g., fine-scale details cannot be preserved.  While most painterly styles involve greater degrees of abstraction, this style illustrates the superior performance of our sequential patch-based optimization, which can achieve accurate image reproduction.

For this style, we use a single layer ($K=1$), without semantic segmentation, and run four 
coarse-to-fine painting passes $(P=4)$. The number of strokes $s$ per patch increases with each layer, following the formula $s_p = (p+5)^2$. Stroke thicknesses are halved across consecutive passes, given by the formula: $ a_p = 2^{1-p} a_1 $, where $a_p$ and $a_1$ are the $p$-th and thicknesses. This approach ensures that the painting captures finer details, rendering a realistic appearance. Paintings employing this methodology are showcased in \Cref{fig:teaser} (c), \Cref{fig:c2f}, \Cref{fig:painterly_vs_realistic} (b), \Cref{fig:control} (c) top, and \Cref{fig:sota_comp_1} (fourth row). We use perceptual losses along with pixel losses to capture higher frequency areas. That is, we optimize the stroke parameters following \Cref{eq:loss}, where $\alpha=1$ and $\beta=0.01$.
This style achieves superior reconstruction details than previous methods as seen in \Cref{sec:qual}, and in \Cref{fig:teaser} (c), and \Cref{fig:sota_comp_1} (fourth row). Analyzing these paintings, high-frequency areas are captured with higher definition than previous methods, that is, strokes follow edges and shapes more accurately, resulting in higher-fidelity paintings. 


\paragraph{Painterly Style.}
The goal of this style is to generate a more loose representation of the input image, yet capturing enough details so that the original scene is represented. Leveraging semantic segmentation, our approach mirrors the human painting process, working
piece-by-piece, focusing on one region or aspect of a painting at a time, and generating different painterly styles. Instead of a uniformly detailed canvas, semantic control offers refined governance over stroke distribution, tailored to distinct semantic zones. Our dynamic attention maps, inspired by visual working memory, are configured to a selective mode, honing in on areas necessitating enhanced detail. The stroke parameters are optimized solely based on pixel loss, that is, we set $\beta=0$ in \Cref{eq:loss}.
For each painting pass $p$, we found enough to have a stroke budget of 16 strokes per patch, which yields a distinctive painterly style. This approach prioritizes artistic expression over mere replication, yielding a rendition that captures the essence rather than the exact likeness of the input image, as demonstrated in \Cref{fig:teaser} (b), \Cref{fig:biases}, \Cref{fig:painterly_vs_realistic} (a), \Cref{fig:control} (a, bottom), (b, top), and \Cref{fig:sota_comp_2} (b,c). In these figures, we can see that this style has the loose quality of typical impressionistic paintings, yet forms are defined and the subject matter is clear.



\paragraph{Semantic Abstraction.}
Unlike the previous painterly style, which still prioritizes a good balance between accuracy and expression, this style leans with a heavier emphasis on expression. By adjusting the optimization procedure over painting passes $P$ and semantic layers $K$, we can generate abstract representations, offering a novel painting synthesis. To do this, the user associates different optimization parameters such as $P$, stroke budget, and number of selective dynamic attention maps $V$ to different semantic layers $k$. Since this style does not focus on high-fidelity, stroke parameters are optimized based on $L_1$ loss, setting $\beta = 0$ in \Cref{eq:loss}. This generates organic and abstract representations of the object of interest, focusing on expression over representation, and can be applied universally or selectively to accentuate specific semantic zones. See how in \Cref{fig:teaser} (a), and \Cref{fig:sota_comp_2} (a), the abstract paintings loosely define the subject matter, prioritizing heavy abstract expressionism, in contrast to the painterly style, which captures the subject matter, but still leaving room to the viewer's interpretation. A practical illustration: buildings can be rendered in finer detail to spotlight their significance, whereas adjacent elements like streets, pedestrians, or foliage are abstracted, adding an artistic bias to the scene (refer to \Cref{fig:biases} and \Cref{fig:control}). As a result, the painting looks more intentional, organic and fluid, similar to how human artists paint. \Cref{fig:teaser} (a) and \Cref{fig:sota_comp_2} (a) are excellent examples of the human-like qualities of this style, with brushwork being loose and gestural, creating a rich tapestry of dynamic shapes. Other examples using this style are shown in \Cref{fig:sota_comp_1} (bottom row), where semantic layers are entirely abstracted from their context. For further details on the parametrization of this style, please refer to supplementary material.

\subsection{Comparison with State-of-the-Art}
\label{sec:exp-sota}
\input{tables/attributes}

We evaluate our work against three notable parametric methods: an optimization-based method ``Stylized Neural Painting" (SNP) \cite{Zou_2021_CVPR}, a learning-based method with Transformer ``Paint Transformer" (PT) \cite{Liu_2021_ICCV}, and a RL method ``Learning to Paint" (L2P) \cite{huang2019learning}. Our method stands out by supporting any resolution or aspect ratio, preserving artwork quality, unlike L2P and PT which resize outputs, often reducing quality.
Moreover, SNP is limited to square formats, restricting its application. An overview of algorithmic attributes is shown in \Cref{tab:algorithms_attributes}. While all methods can output paintings at high resolutions, L2P's results are blurry and seams are visible. For a fair comparison, all paintings have been generated using the same number of strokes. 

\paragraph{Qualitative Comparison}
\label{sec:qual}
\input{tables/new_user_study}

\input{figures/fig_tex/grid_sota_p1_small}
\input{figures/fig_tex/grid_sota_p2}


A visual comparison of our method against previous methods is depicted in \Cref{fig:teaser}, \Cref{fig:sota_comp_1} and \Cref{fig:sota_comp_2}. For more comparisons at larger scale, refer to supplemental. \Cref{fig:teaser} shows how SNP and PT generate impressionistic-like paintings, although the brushwork is patchy and rigid, and both methods are unable to capture edges. This is further seen in \Cref{fig:sota_comp_1}, where PT introduces highly contrasted brush sizes for different semantic regions, and noise in high-frequency areas. SNP's rigid and uniform patches do not capture well finer details. Out of the previous methods, L2P captures micro and macro structures better, in a realistic style, although at large resolutions produces blurry results, and it is unable to generate other painting styles. This is further observed in \Cref{fig:heatmaps}, where L2P cannot distinguish between semantic areas or objects, treating the entire canvas with the same emphasis. 

\paragraph{User Study.} We assessed our painting method's effectiveness through a user study involving 30 participants, contrasting our outputs with three existing techniques. Participants judged the paintings based on realism and visual appeal across six painting sets, for a total of 12 questions, using a two-way comparison, and collecting 360 responses. As shown in \Cref{tab:user_study}, our method is preferred over existing methods in all tasks. Overall, our method substantially outperforms previous methods in realism, selected by $92.42\%$ of the users. Our method is also preferred in the visual appeal task, voted by $68.94\%$. When evaluating our method on visual appeal, our strongest competitor is L2P, selected by $43.18\%$ of the participants.





\paragraph{Quantitative Comparison}
\input{figures/fig_tex/heatmaps}

\input{tables/quant}
To further substantiate our claims, we compare the performance of our model on reconstruction of the original input against previous methods, and use pixel loss $L_1$ and perceptual loss $L_{perc}$ as evaluation metrics, reported in \Cref{tab:quant}. We select 10 images from each domain, resize them to 512x512 and set the number of strokes to 4000. Our method outperforms previous methods in high-fidelity realistic paintings across all evaluated domains, which is consistent with the qualitative results obtained in our user study. 

To visually quantify how each method places strokes on the canvas, \Cref{fig:heatmaps} compares the stroke distributions for the middle column paintings shown in \Cref{fig:sota_comp_1} and \Cref{fig:sota_comp_2}. Our method (d) shows a more intentional stroke placement and clarity in stroke application. In contrast, SNP (a) shows a scattered, uniform treatment of scene elements, while our approach strategically focuses on two key areas, corresponding to the vegetation, enhancing the composition's structure. L2P (b) applies strokes uniformly in a scene-agnostic way, lacking the focal emphasis of our method. PT (c) creates abrupt density transitions due to its multi-scale approach, unsuitable for gradual tonal shifts, and fails to disentangle high-density areas, not differentiating between scene elements like trees and roads. Our method, however, clearly delineates regions with varying stroke densities, offering a nuanced and intentional subject representation. See supplemental for additional details.

\paragraph{Efficiency} We compare the efficiency of the different previous methods in a single RTX A6000 Nvidia GPU for a single image (see \Cref{tab:eff}). While neural methods require training, they are faster at inference time than optimization methods. However, when comparing with previous optimization methods \cite{Zou_2021_CVPR}, our method is significantly faster (4x) for the same number of strokes (1600) and resolution size (1024x1024). This is specially important because optimization methods require an optimization process per image. For a fair comparison, we use the uniform setting for our dynamic maps, and set the style to realism.

%% file: figures/fig_tex/coarse2fine.tex
\begin{figure}[ht]
\begin{center}
\includegraphics[width=\linewidth]{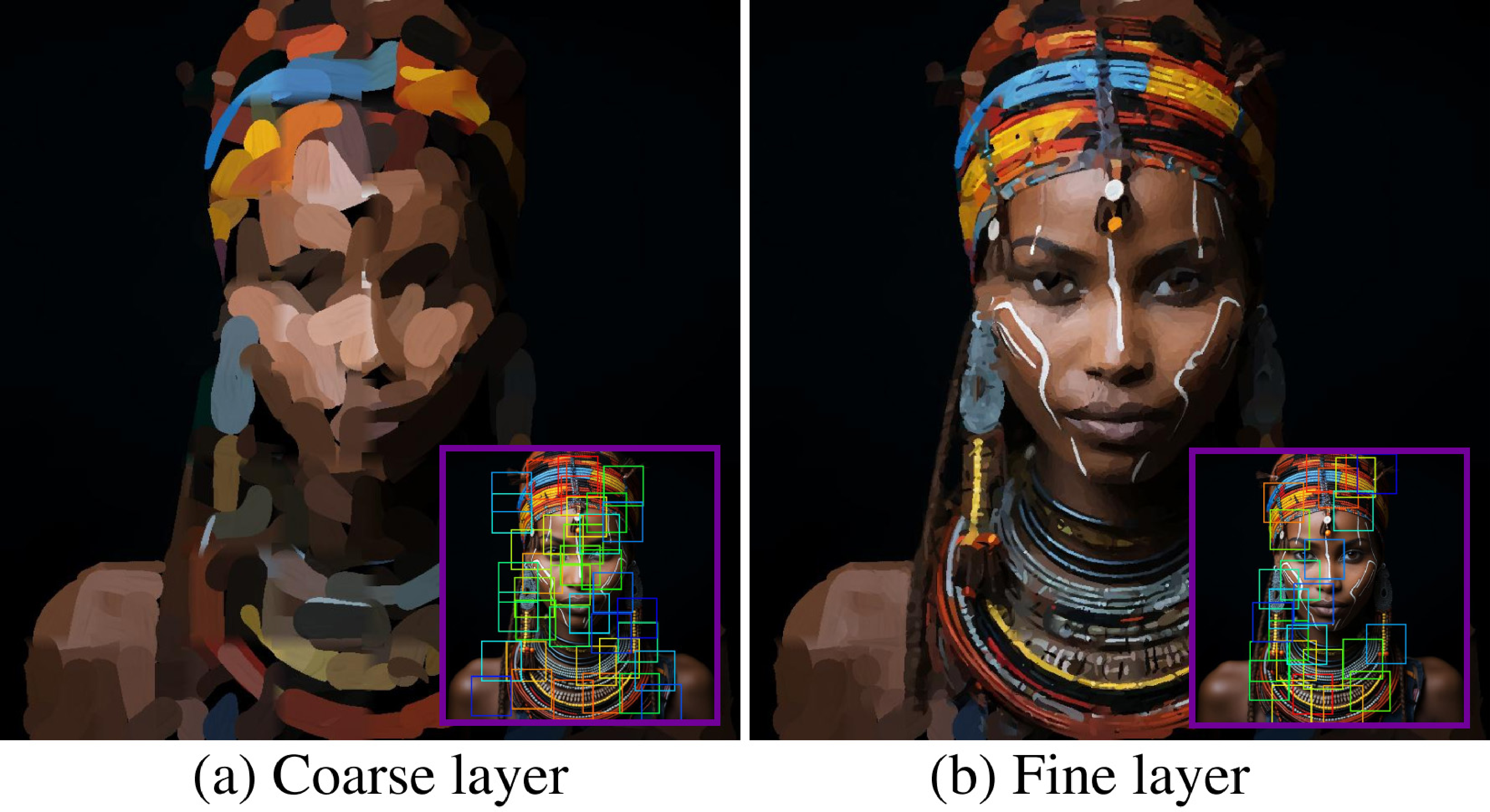}
\end{center}
\vspace{-1em}
\caption{Coarse-to-fine approach for a painterly portrait of a woman. (a) Initial coarse layer lays foundational strokes for a later refinement (b), guided by dynamic attention maps (insets).}
\label{fig:c2f}
\end{figure}

%% file: figures/fig_tex/biases.tex
\begin{figure}[t]
\begin{center}
\includegraphics[width=\linewidth]{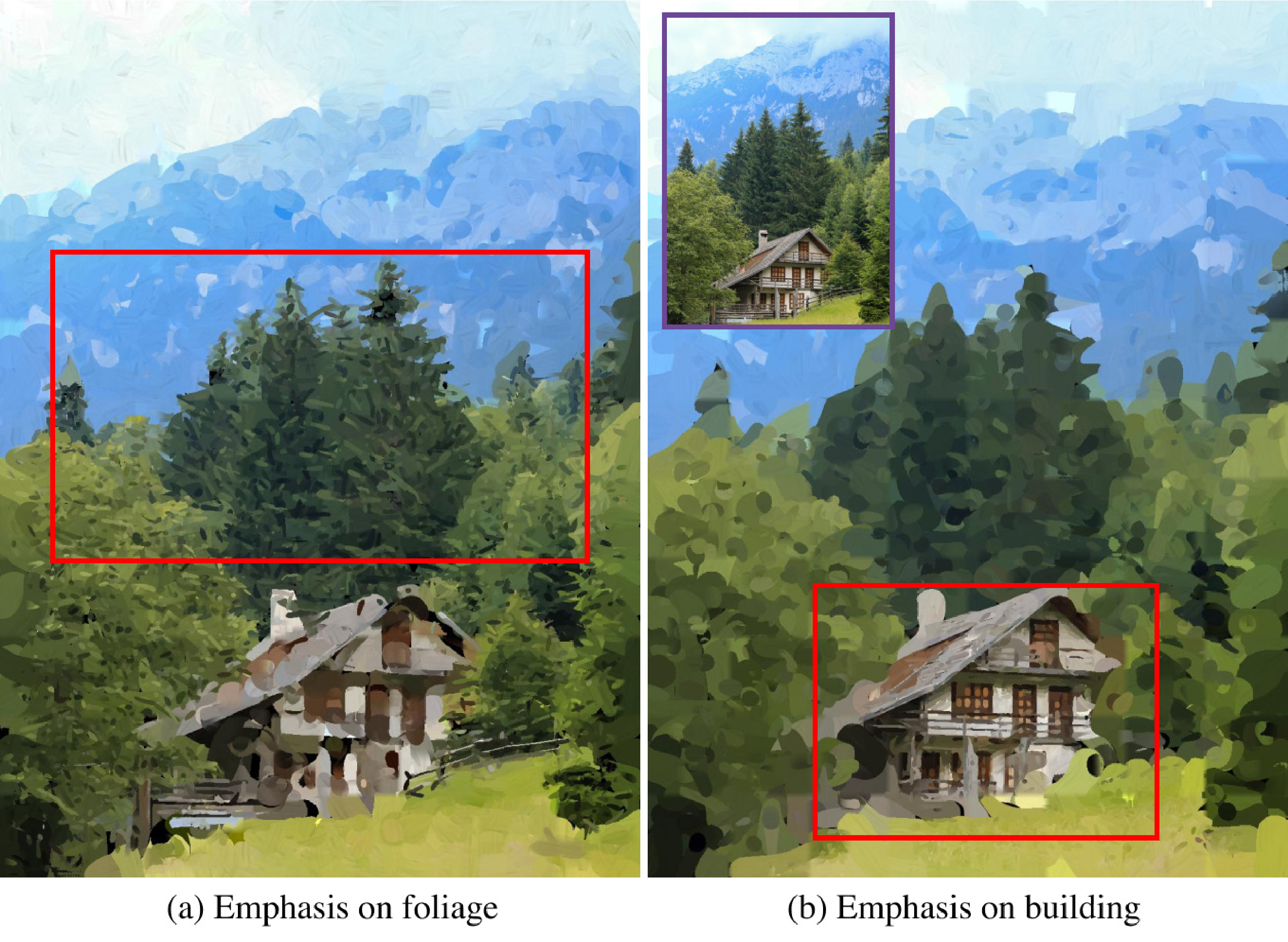}
\end{center}
\vspace{-1em}
\caption{Our approach offers enhanced precision over the canvas's semantic regions, directing specific emphases in the artwork: (a) accentuates the foliage, capturing its essence, while (b) concentrates on highlighting the architectural details of the building.}
\label{fig:biases}
\end{figure}

%% file: figures/fig_tex/painterly_vs_realistic.tex
\begin{figure*}[htb!]
\begin{center}
\includegraphics[width=\textwidth]{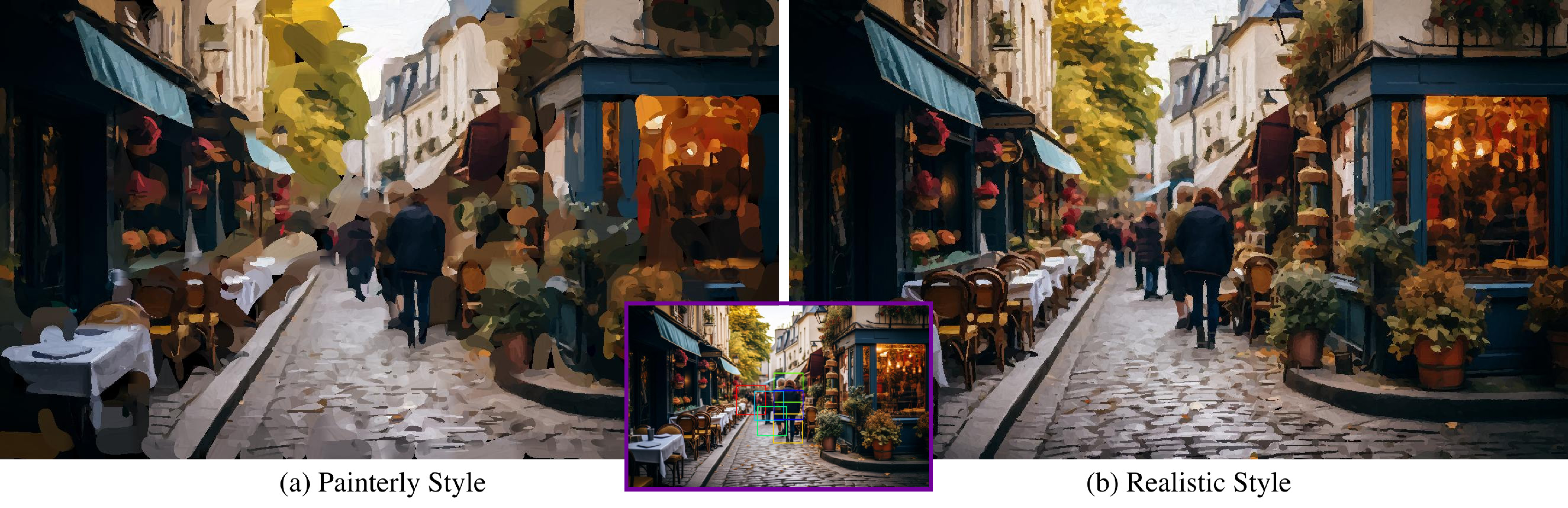}
\end{center}
\vspace{-1em}
\caption{Different painterly styles given the same input. (a) Painterly style achieved by the use of our segmentation pipeline and selective dynamic attention maps. (b) Realistic style achieved by using uniform dynamic maps.}
\label{fig:painterly_vs_realistic}
\vspace{-1em}
\end{figure*}

%% file: figures/fig_tex/control.tex
\begin{figure}[t]
\begin{center}
\includegraphics[width=\linewidth]{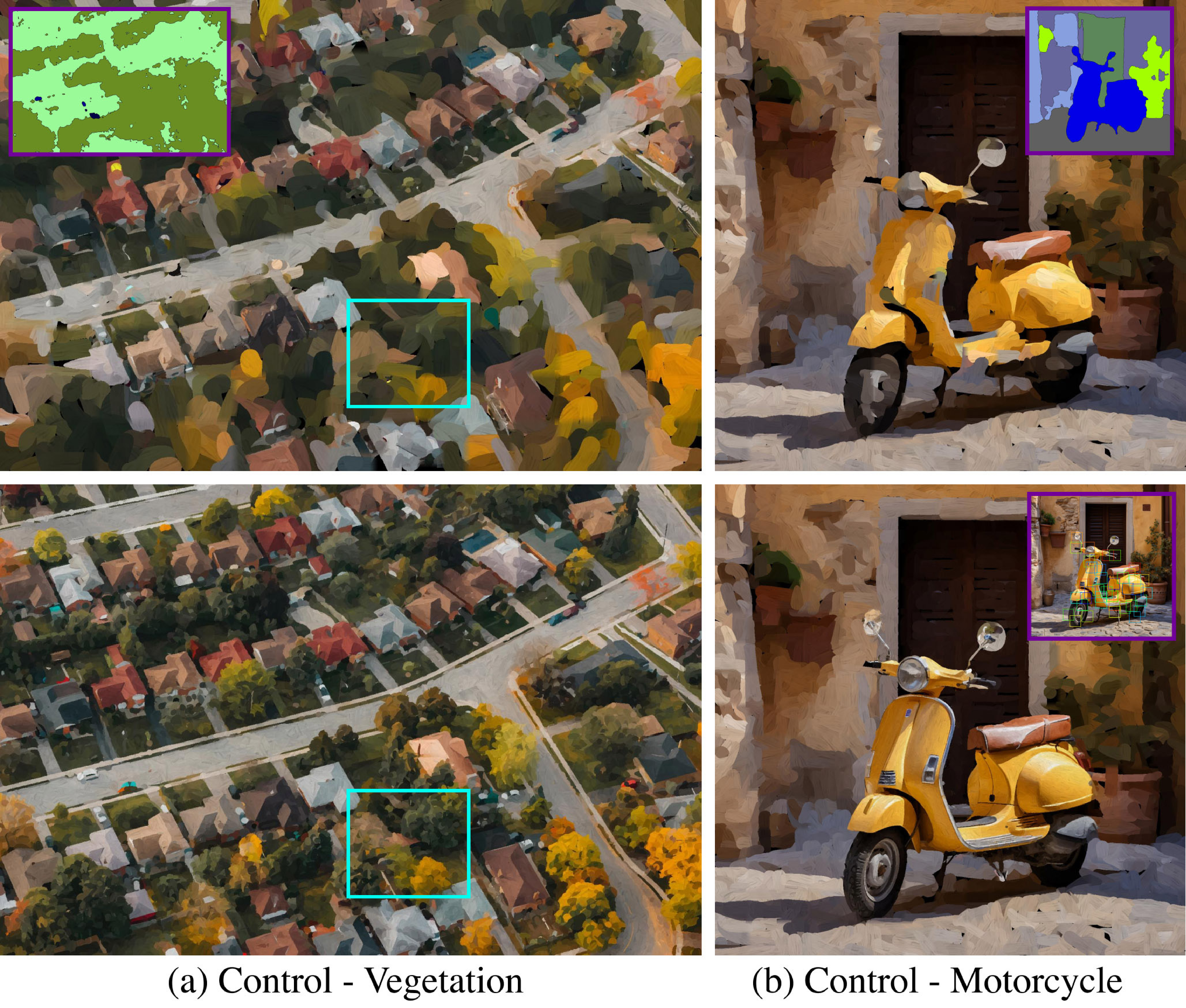}
\end{center}
\vspace{-1em}
\caption{Demonstration of control over the painting. (a) The painting focal point falls in the buildings while leaving the vegetation abstracted. (b) The motorcycle is emphasized by adding more detail than the background, which is coarsely painted.}
\label{fig:control}
\vspace{-1em}
\end{figure}

%% file: tables/attributes.tex
\begin{table}
    \centering
    \begin{adjustbox}{width=\linewidth, center}
    \begin{tabular}{ccccc}
    \toprule
           & Control & Style Var. & Large-Scale & High-low Freq. \\
    \midrule
        PT & \text{\ding{55}} &  \text{\ding{55}}  & \checkmark & \text{\ding{55}}\\
        SNP & \text{\ding{55}} & \checkmark & \checkmark & \text{\ding{55}} \\
        L2P & \text{\ding{55}} & \text{\ding{55}} & \text{\ding{55}} & \checkmark \\
        Ours & \checkmark & \checkmark & \checkmark & \checkmark \\
    \bottomrule
    \end{tabular}
    \end{adjustbox}
    \caption{Comparison with previous methods on attributes like painting control, style variations, large-scale parametric painting, and high-low frequency capture.}
    \label{tab:algorithms_attributes}
\vspace{-1em}
\end{table}

%% file: tables/new_user_study.tex
\begin{table}[ht]
\begin{adjustbox}{width=\linewidth, center}
  \centering
  {\scriptsize
  \begin{tabular}{l c r}
    \toprule
    Method & Realism & Visual Appeal \\
    \midrule
    Ours vs. PT \cite{Liu_2021_ICCV} & 81.82\% & 79.55\% \\
    Ours vs. SNP \cite{Zou_2021_CVPR} & 100\% & 70.45\% \\
    Ours vs. L2P \cite{huang2019learning} & 95.45\% & 56.82\% \\
    \bottomrule
  \end{tabular}
  }
  \end{adjustbox}
  \caption{Qualitative evaluation based on user preference. This table shows a 2-way comparison between our method and previous methods. }
  \label{tab:user_study}
\end{table}

%% file: figures/fig_tex/grid_sota_p1_small.tex
\begin{figure}[htp]
\centering

\begin{subfigure}{0.33\linewidth}
\includegraphics[width=\linewidth]{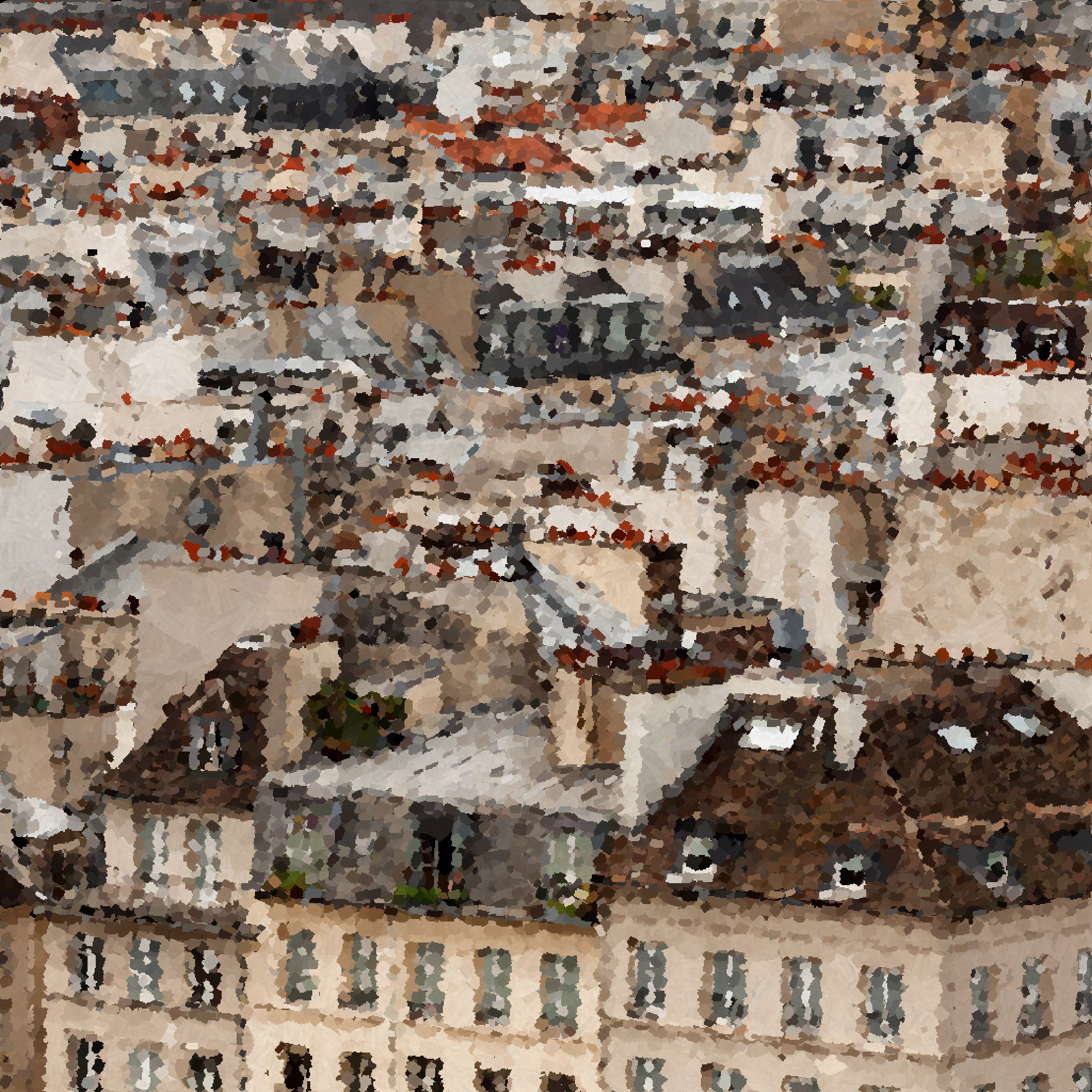}
\caption*{  }
\end{subfigure}\hfill
\begin{subfigure}{0.33\linewidth}
\includegraphics[width=\linewidth]{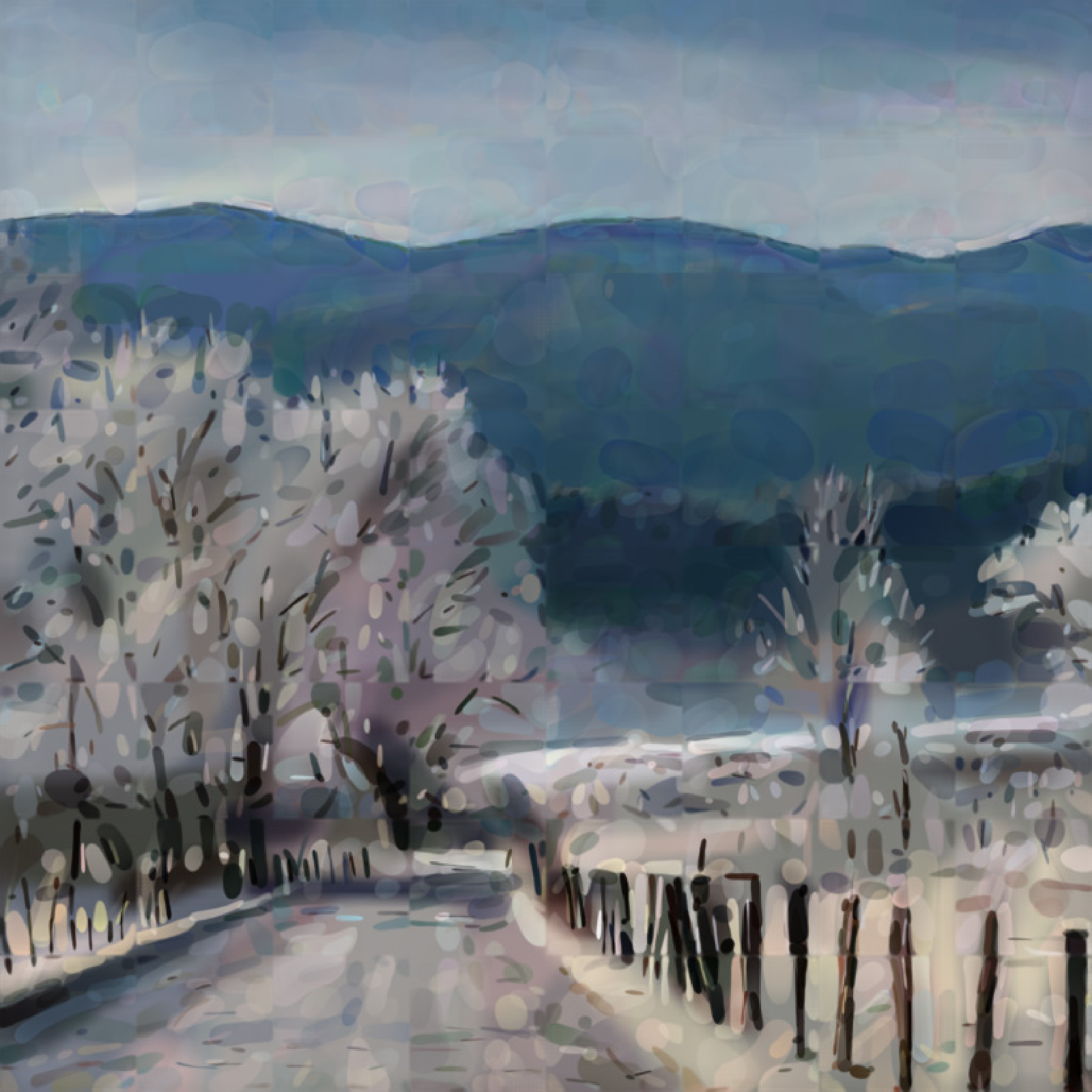}
\caption*{L2P \cite{huang2019learning}}
\end{subfigure}\hfill
\begin{subfigure}{0.33\linewidth}
\includegraphics[width=\linewidth]{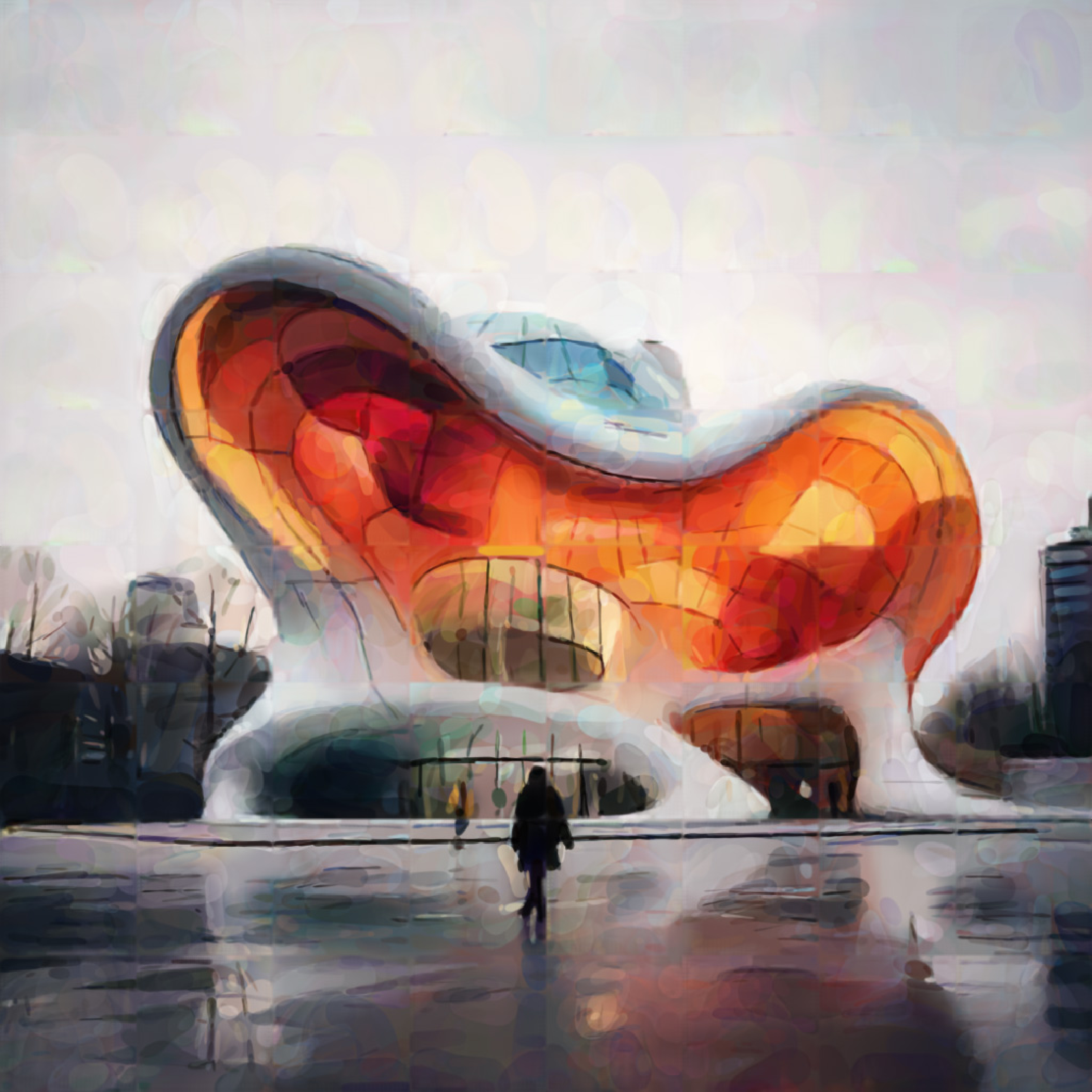}
\caption*{  }
\end{subfigure}

\begin{subfigure}{0.33\linewidth}
\includegraphics[width=\linewidth]{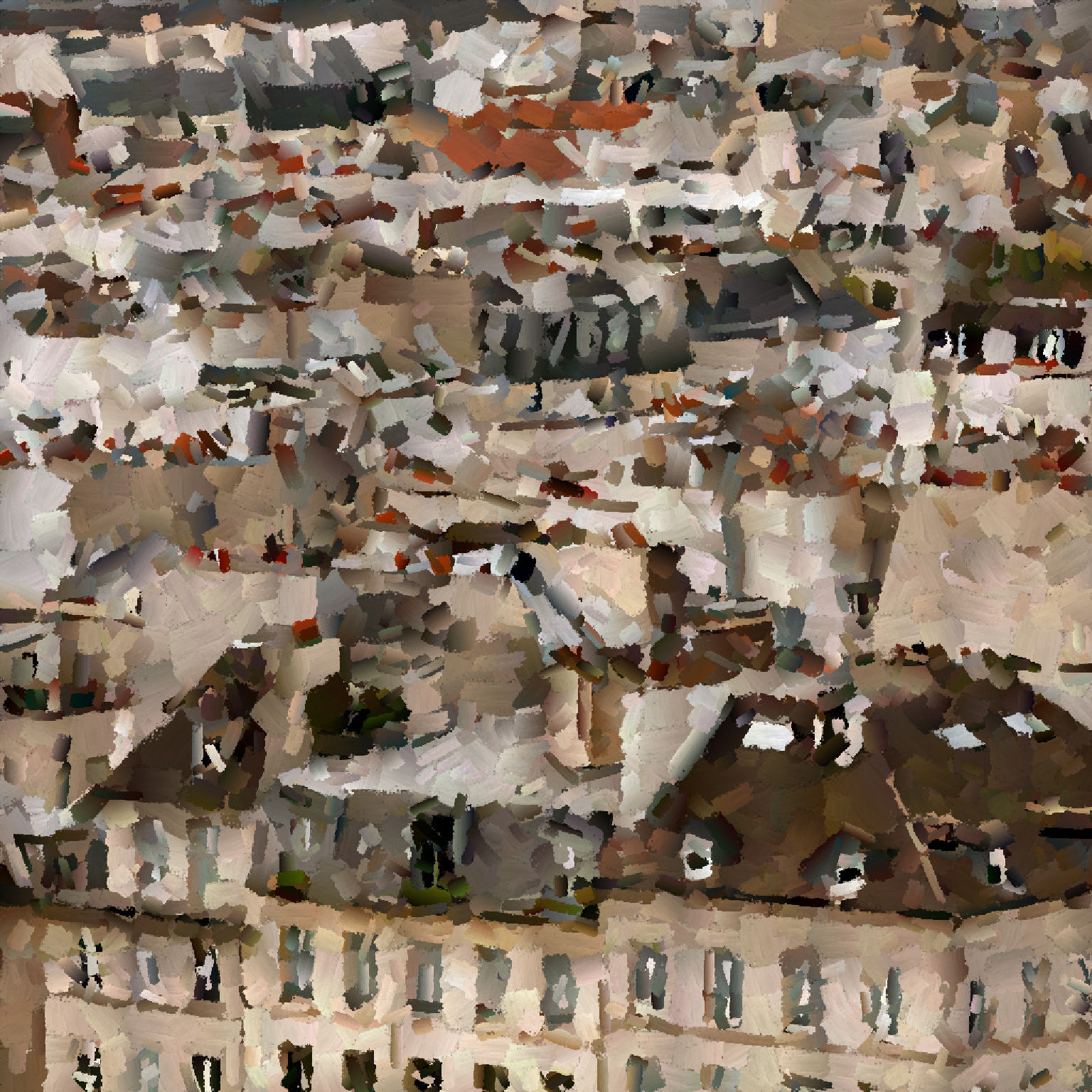}
\caption*{  }
\end{subfigure}\hfill
\begin{subfigure}{0.33\linewidth}
\includegraphics[width=\linewidth]{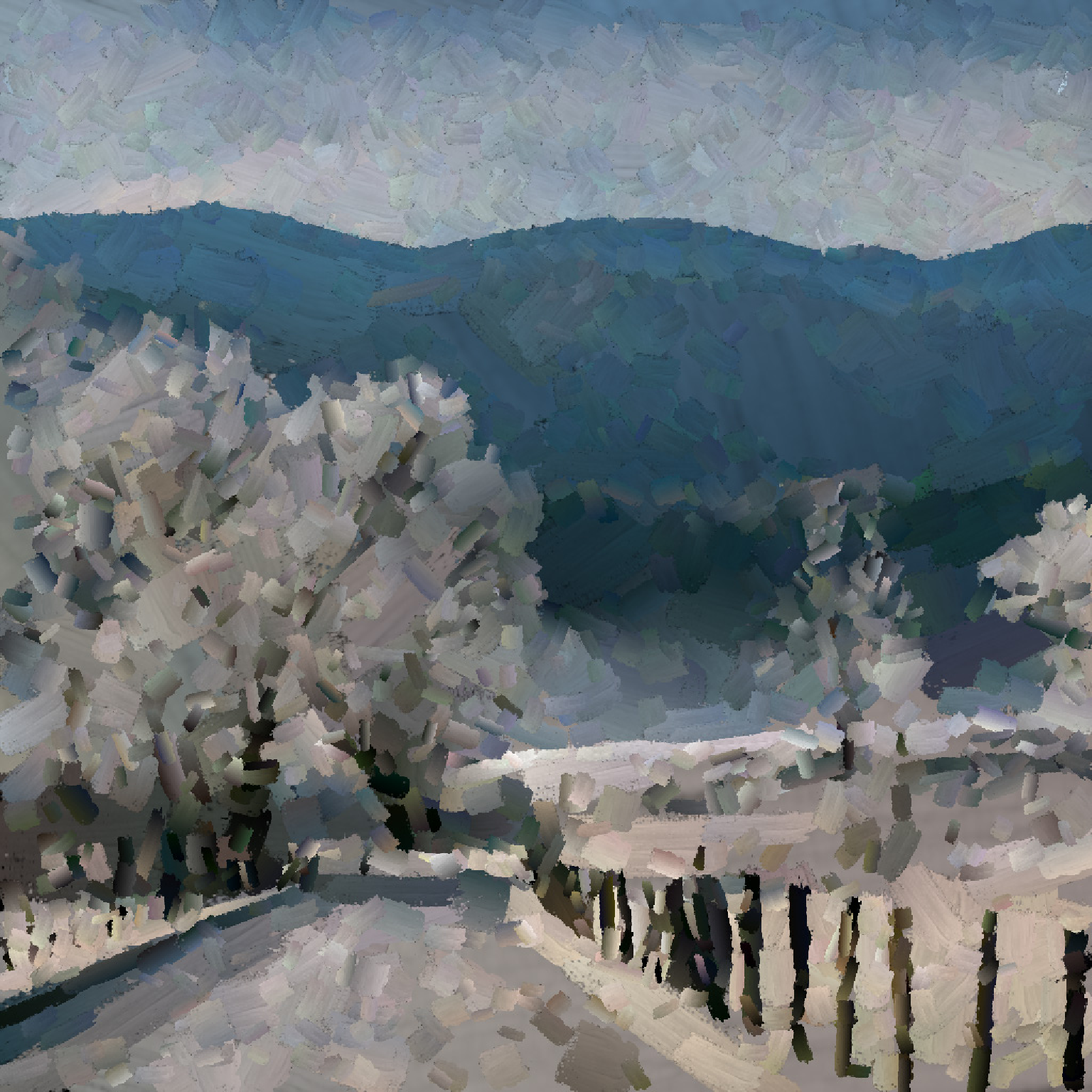}
\caption*{SNP \cite{Zou_2021_CVPR}}
\end{subfigure}\hfill
\begin{subfigure}{0.33\linewidth}
\includegraphics[width=\linewidth]{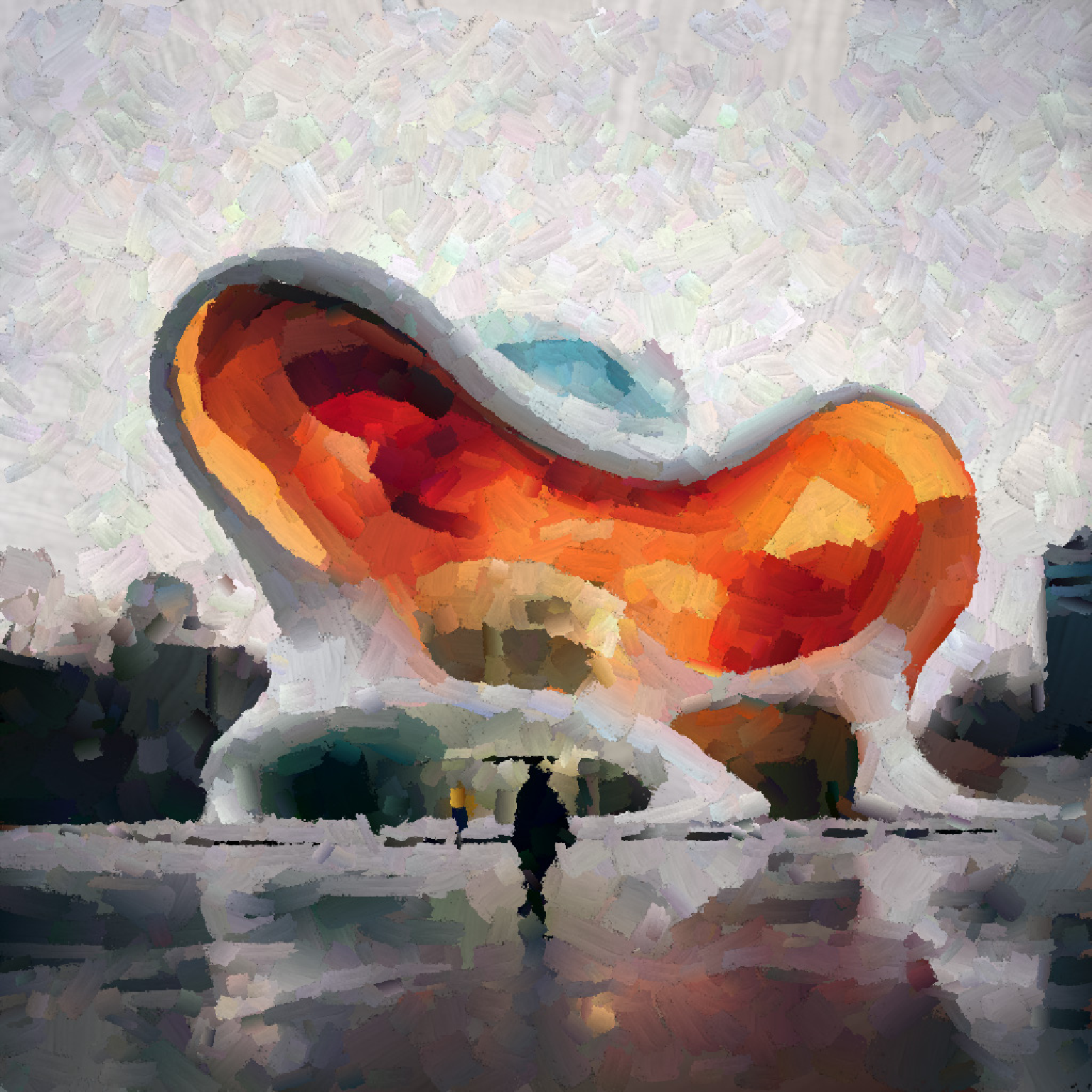}
\caption*{  }
\end{subfigure}

\begin{subfigure}{0.33\linewidth}
\includegraphics[width=\linewidth]{figures/grid_sota/paris_final_pt.eps}
\caption*{  }
\end{subfigure}\hfill
\begin{subfigure}{0.33\linewidth}
\includegraphics[width=\linewidth]{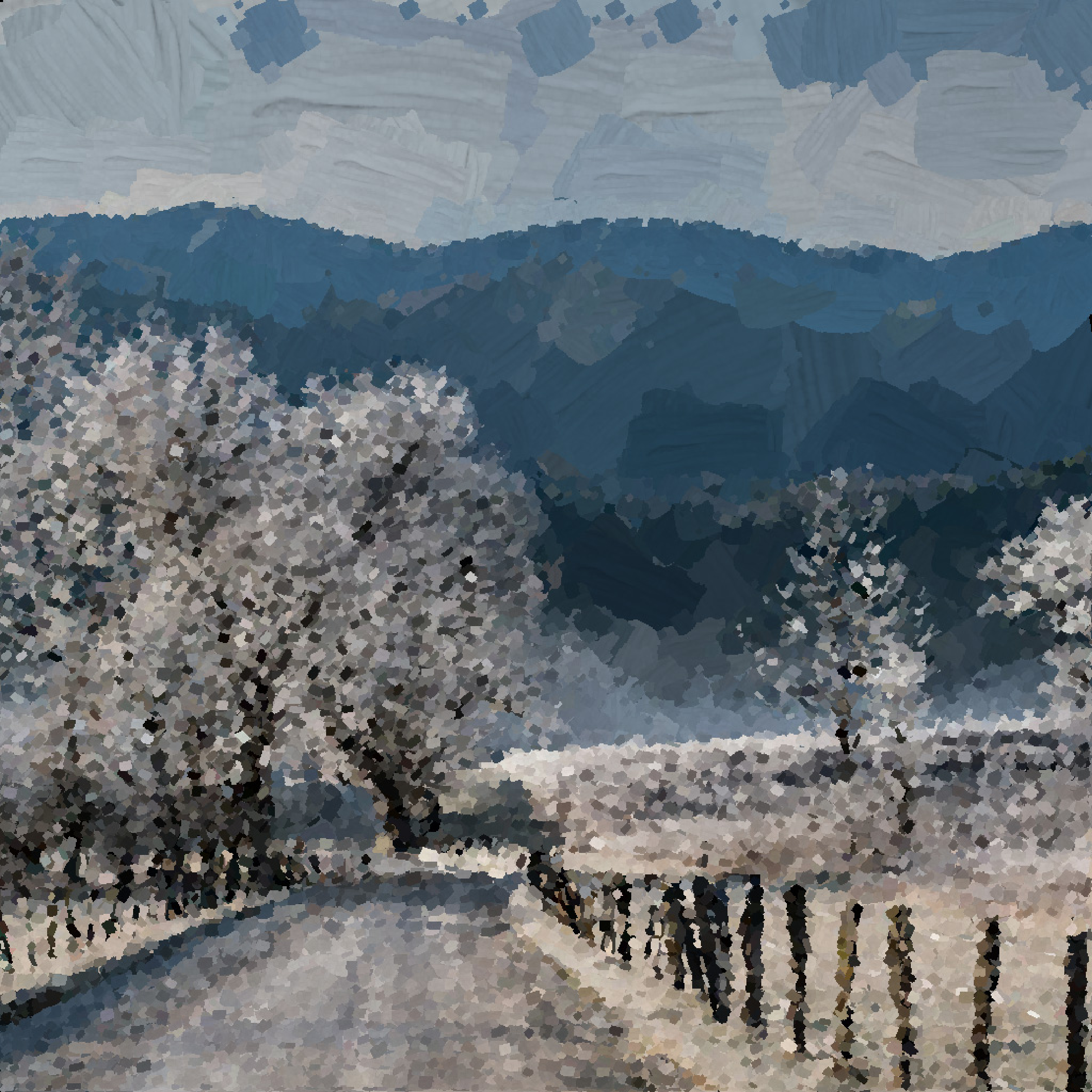}
\caption*{PT \cite{Liu_2021_ICCV}}
\end{subfigure}\hfill
\begin{subfigure}{0.33\linewidth}
\includegraphics[width=\linewidth]{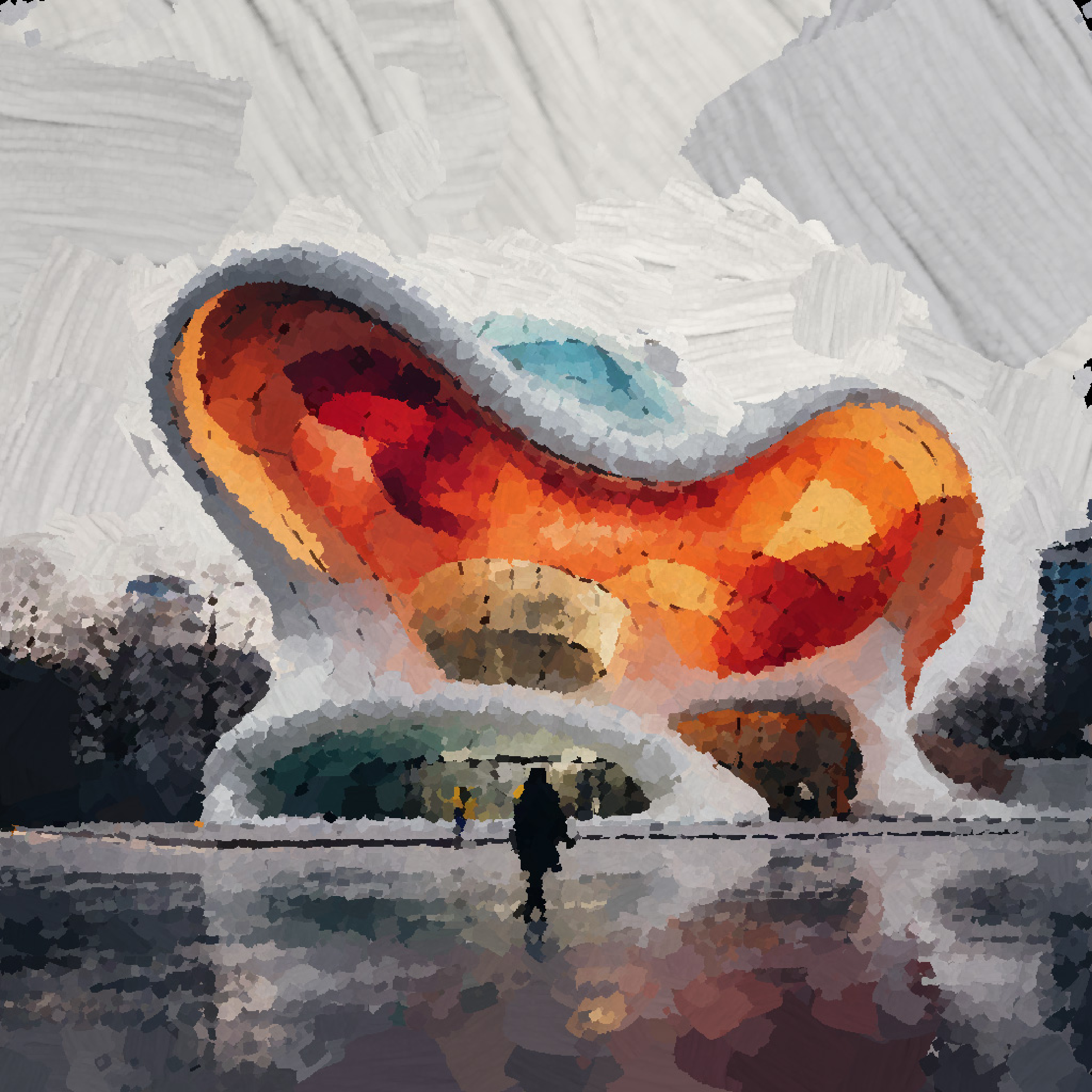}
\caption*{  }
\end{subfigure}

\begin{subfigure}{0.33\linewidth}
\includegraphics[width=\linewidth]{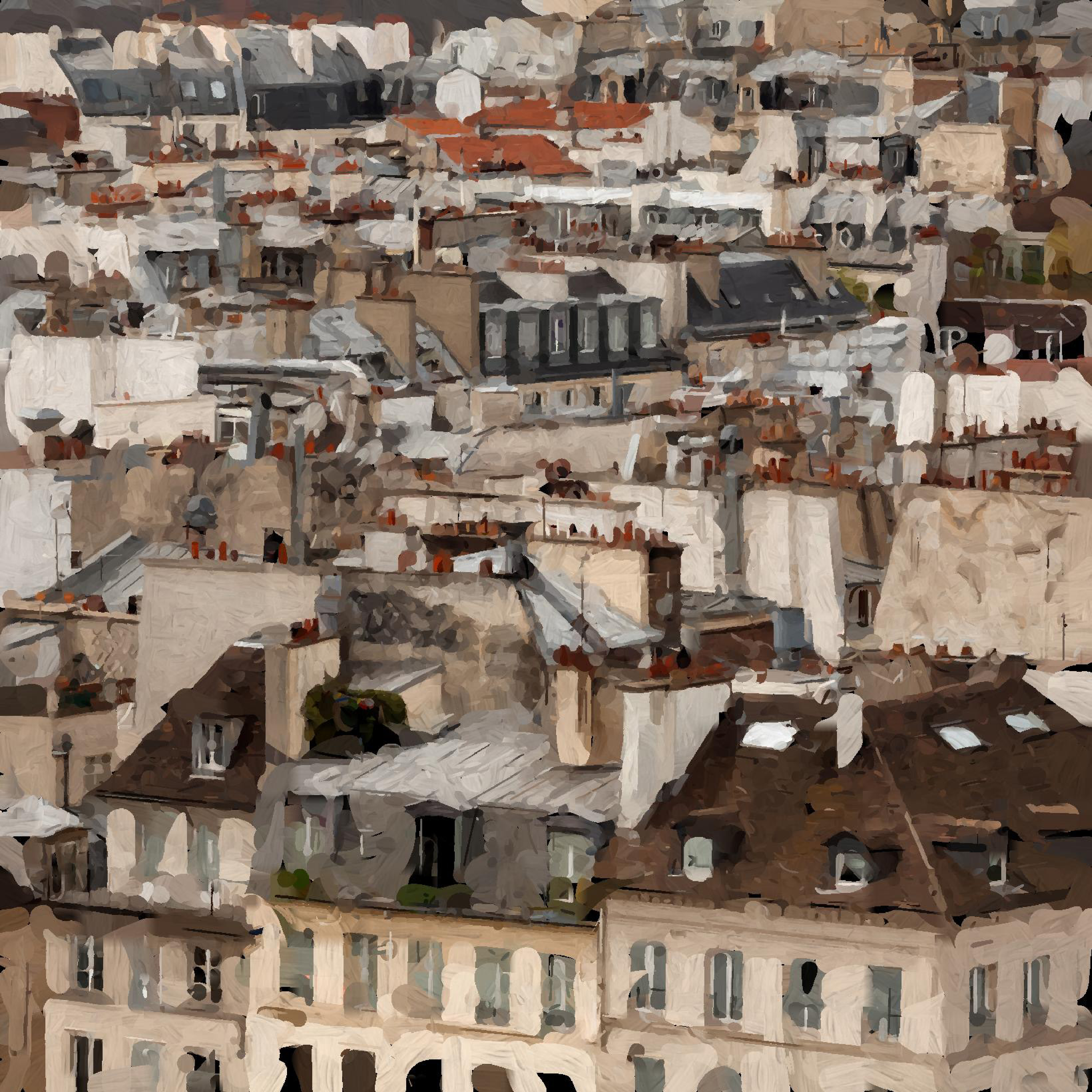}
\caption*{  }
\end{subfigure}\hfill
\begin{subfigure}{0.33\linewidth}
\includegraphics[width=\linewidth]{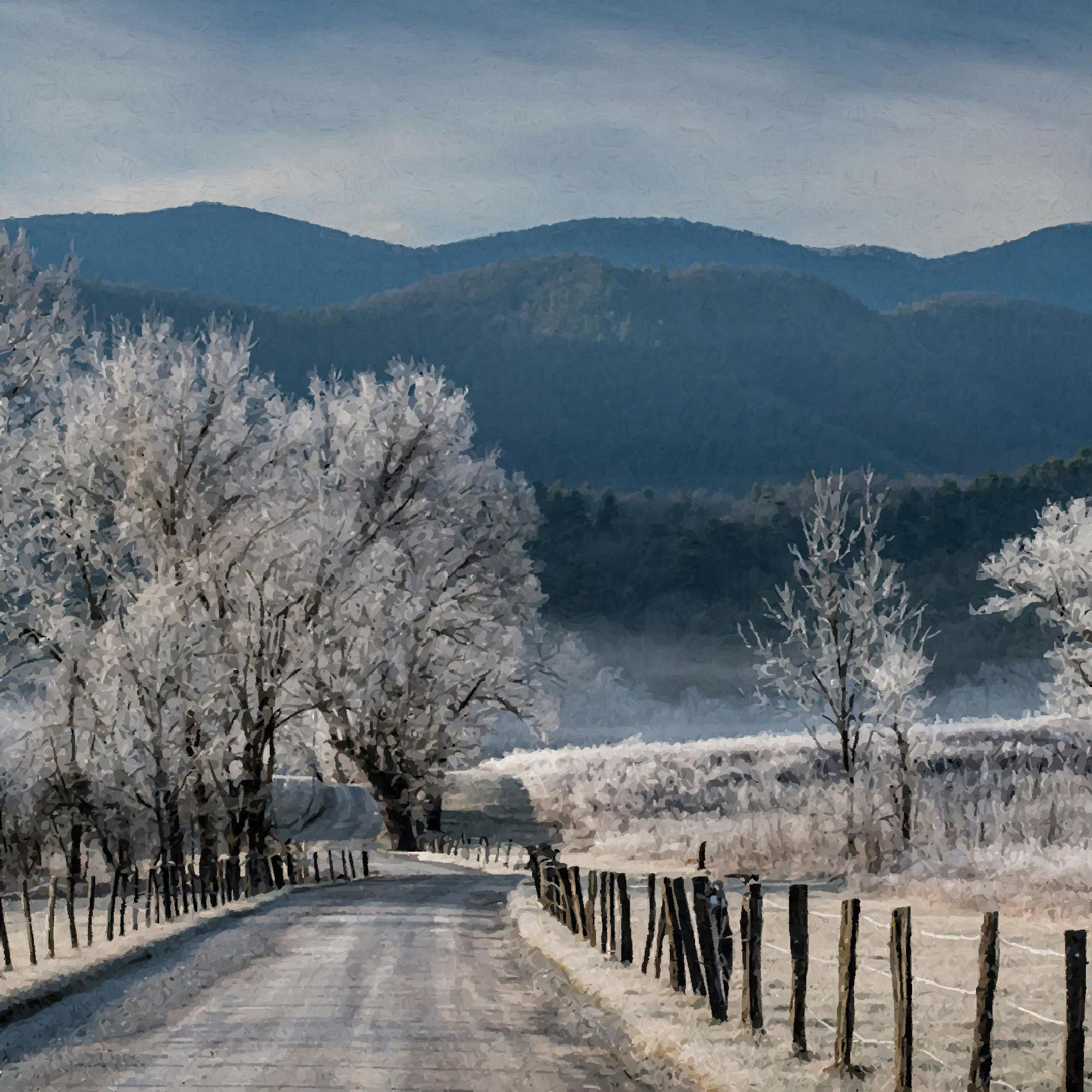}
\caption*{Ours - Realistic}
\end{subfigure}\hfill
\begin{subfigure}{0.33\linewidth}
\includegraphics[width=\linewidth]{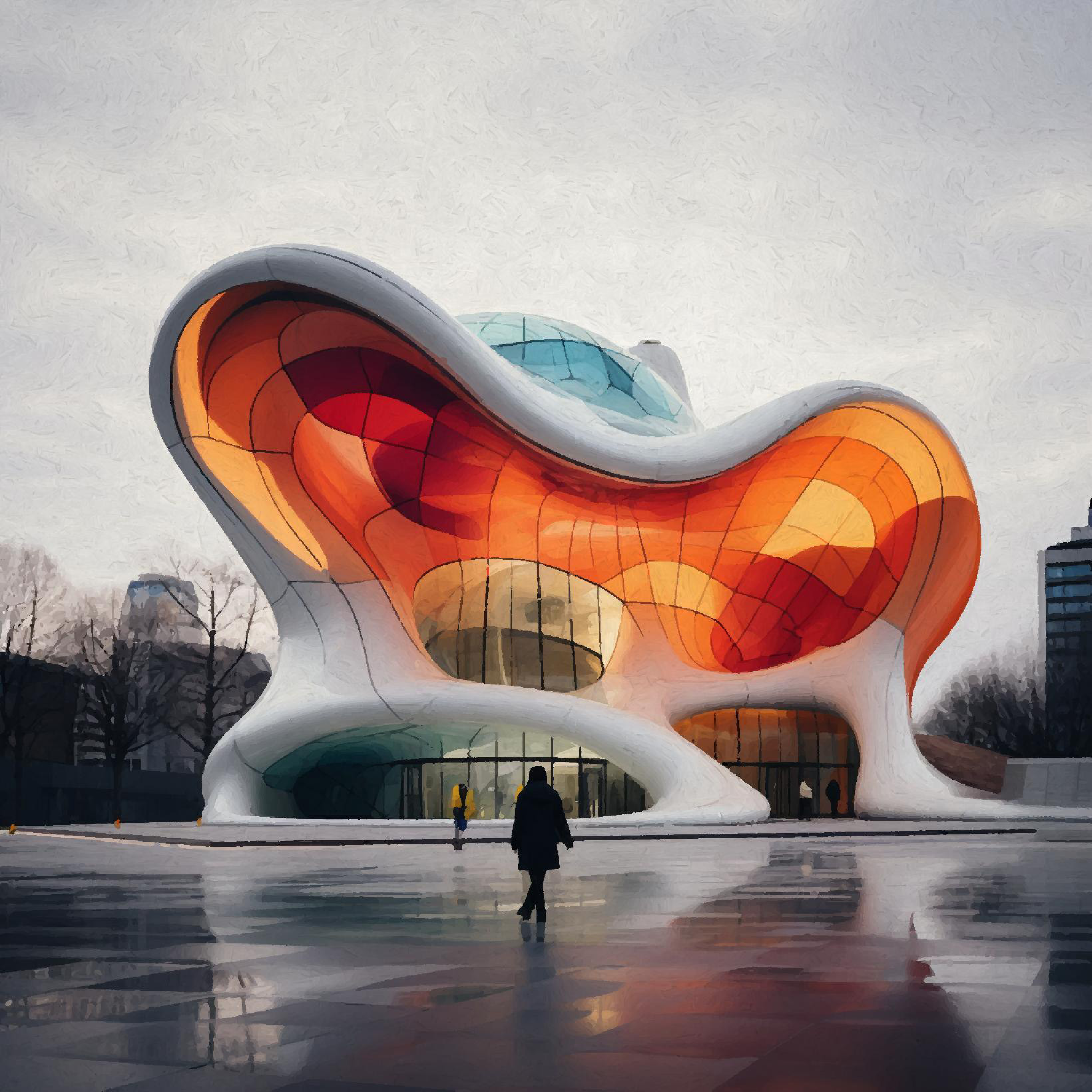}
\caption*{  }
\end{subfigure}

\begin{subfigure}{0.33\linewidth}
\includegraphics[width=\linewidth]{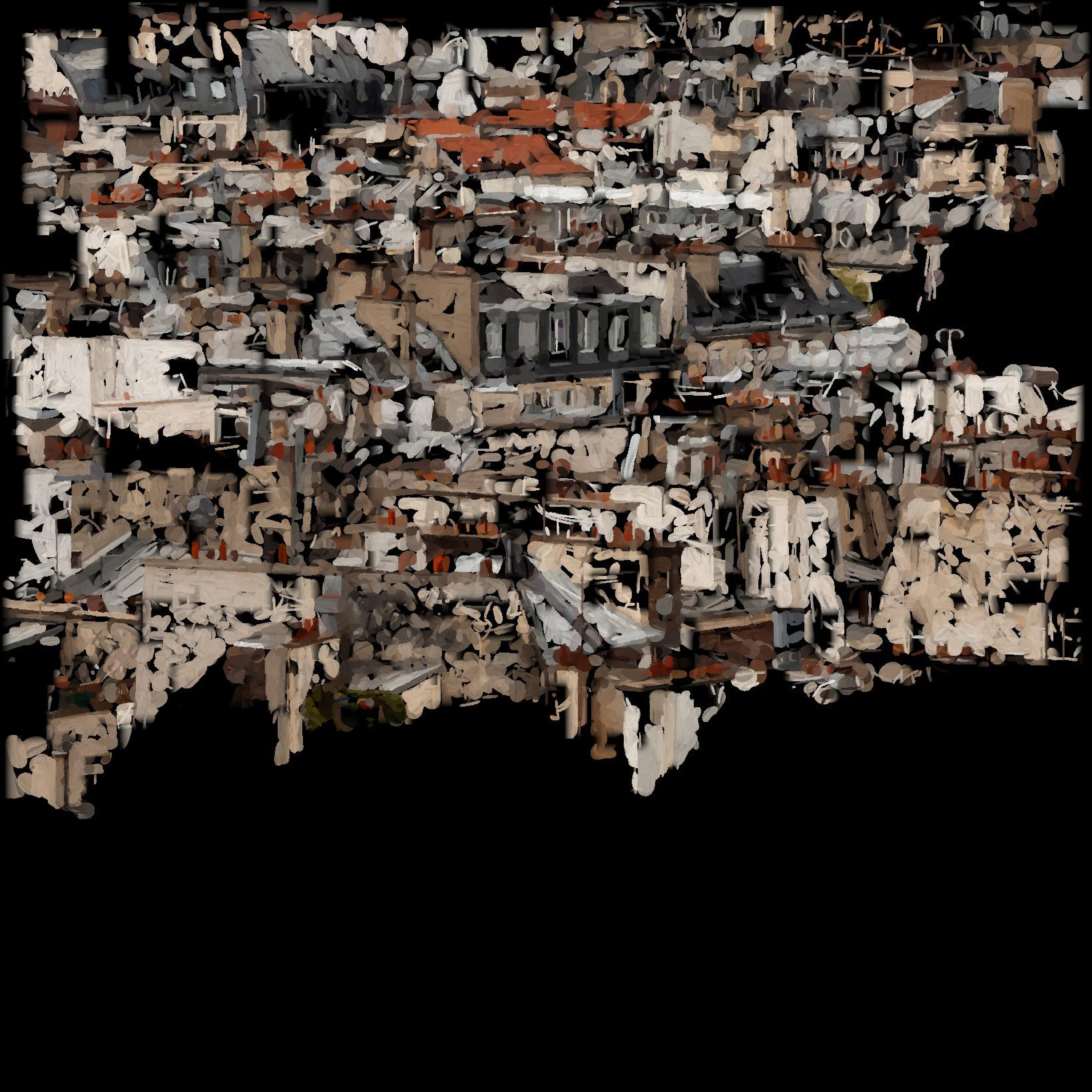}
\caption*{  }
\end{subfigure}\hfill
\begin{subfigure}{0.33\linewidth}
\includegraphics[width=\linewidth]{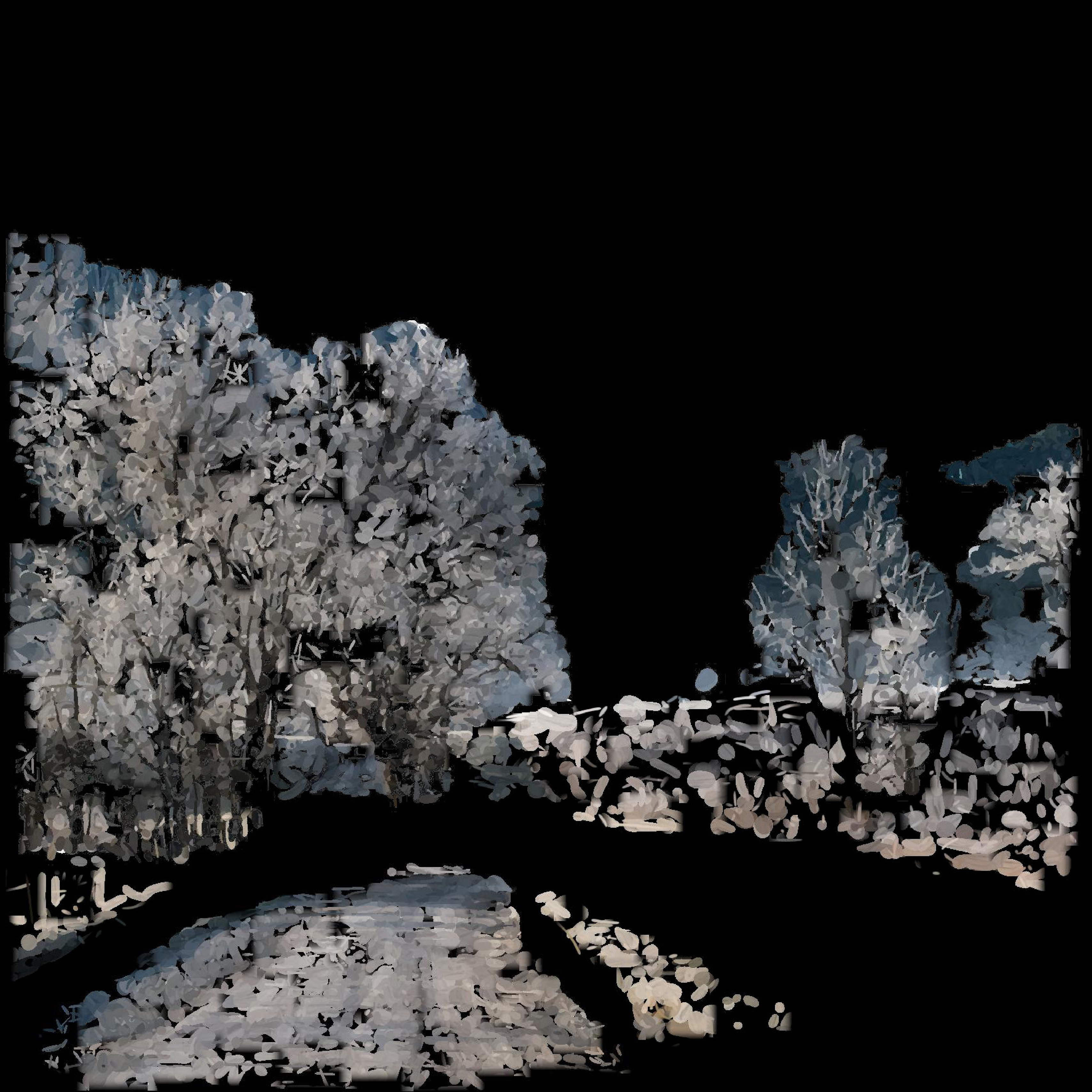}
\caption*{Ours - Abstract}
\end{subfigure}\hfill
\begin{subfigure}{0.33\linewidth}
\includegraphics[width=\linewidth]{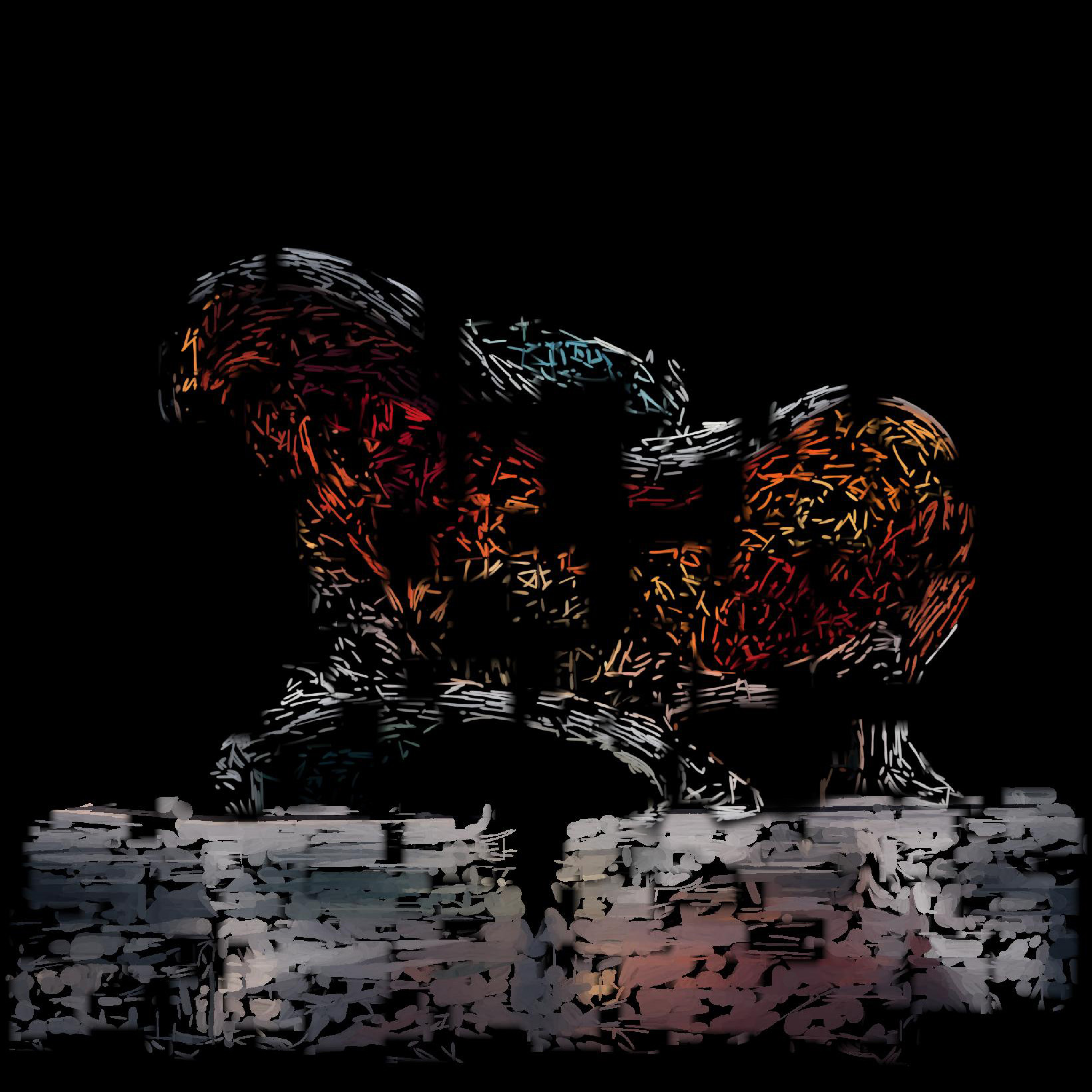}
\caption*{  }
\end{subfigure}

\caption{Comparison of four parametric painting techniques—L2P, SNP, PT, and our approach—across three different scenarios. Our method's outcomes show improved detail rendition and variability. Our realistic approach finely details both high and low-frequency elements, accurately rendering tree branches and window frames. Note L2P's realistic yet blurry style, SNP's broad painterly effect which struggles to capture fine details (tree branches or windows framing), and PT's inconsistent detail and brushstroke scale. Zoom-in recommended for full appreciation of textural nuances.}

\label{fig:sota_comp_1}
\vspace{-1em}
\end{figure}


%% file: figures/fig_tex/grid_sota_p2.tex

\begin{figure*}[htp]
\centering



\begin{subfigure}{0.33\textwidth}
\includegraphics[width=\linewidth]{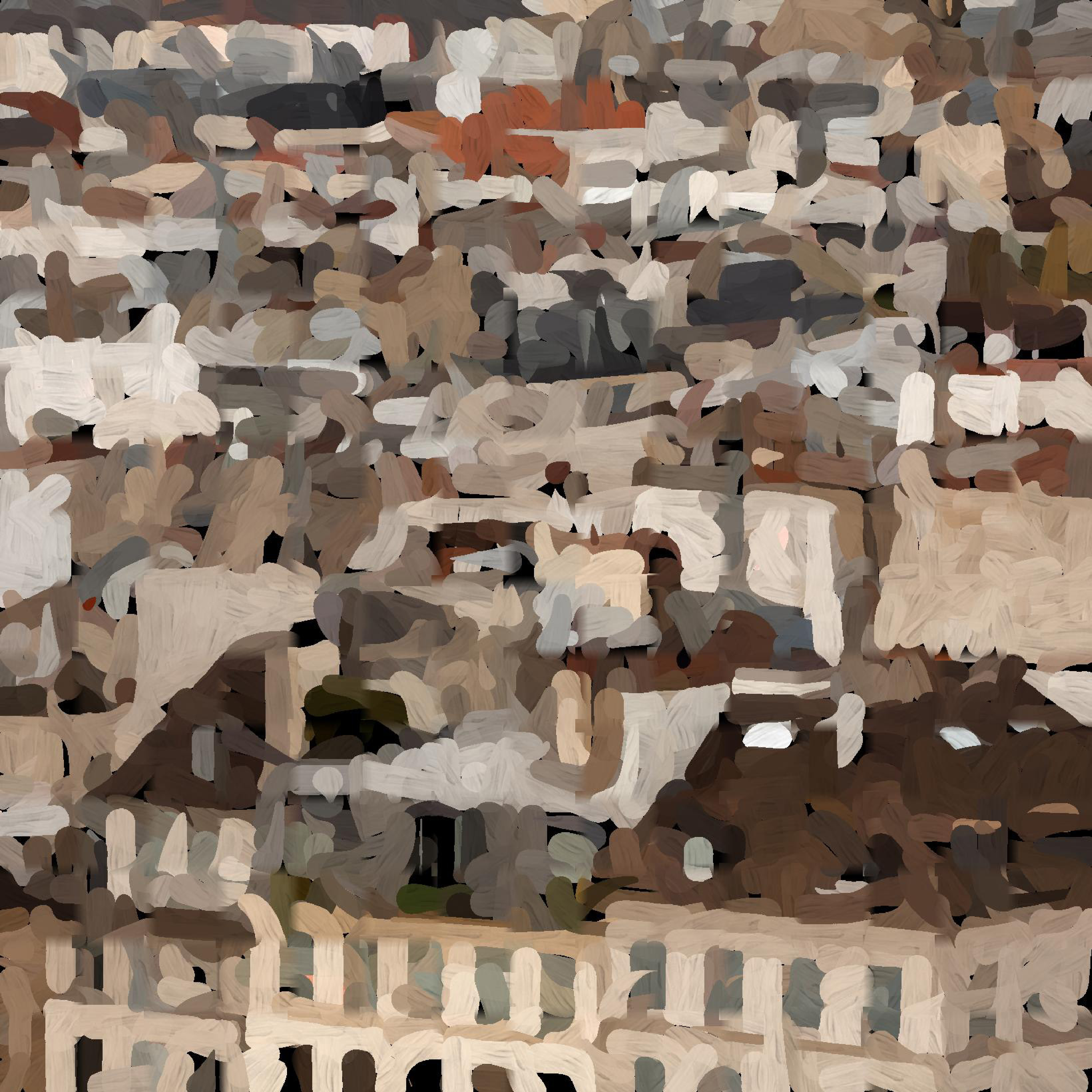}
\caption*{(a) Ours - Abstract}
\end{subfigure}\hfill
\begin{subfigure}{0.33\textwidth}
\includegraphics[width=\linewidth]{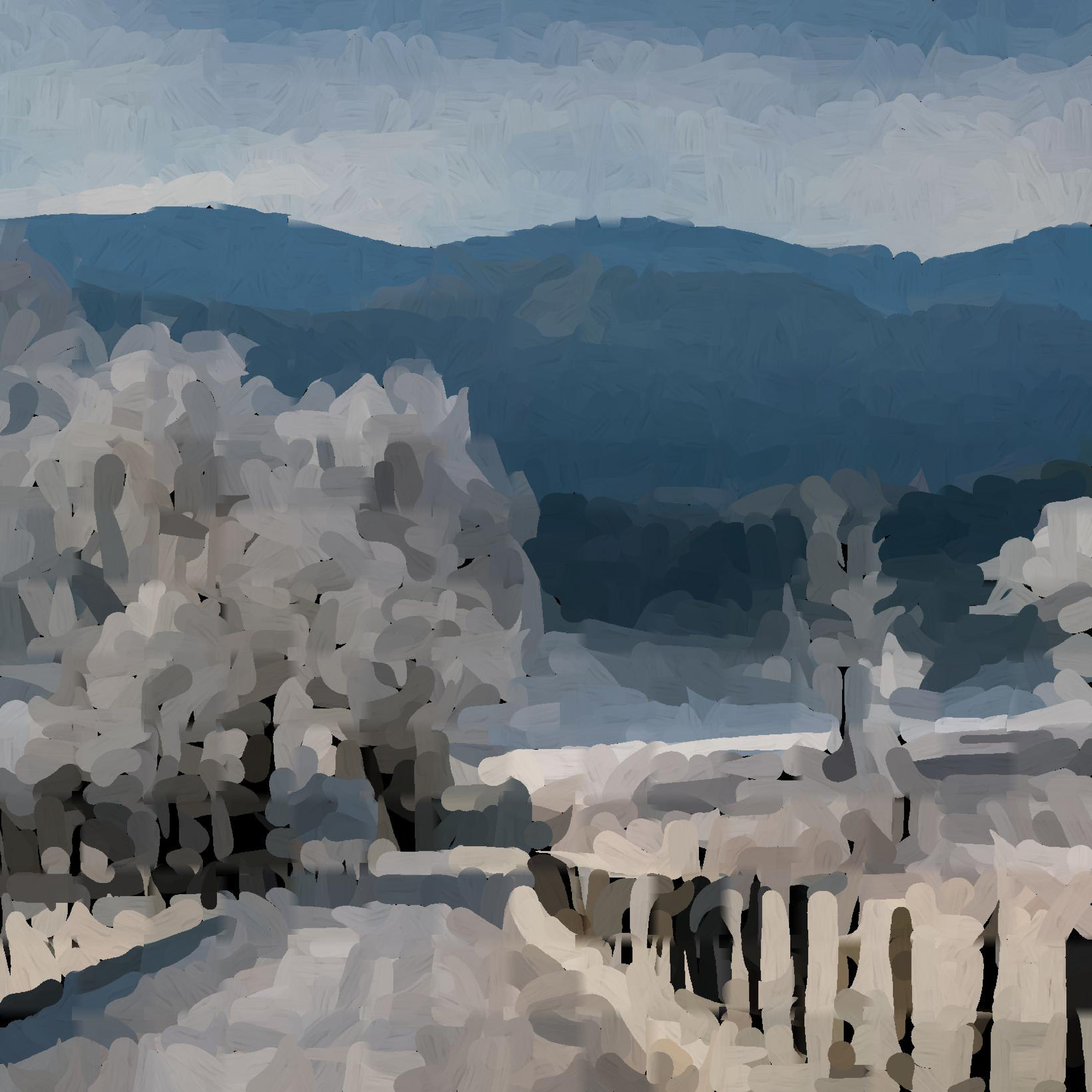}
\caption*{(b) Ours - Painterly}
\end{subfigure}\hfill
\begin{subfigure}{0.33\textwidth}
\includegraphics[width=\linewidth]{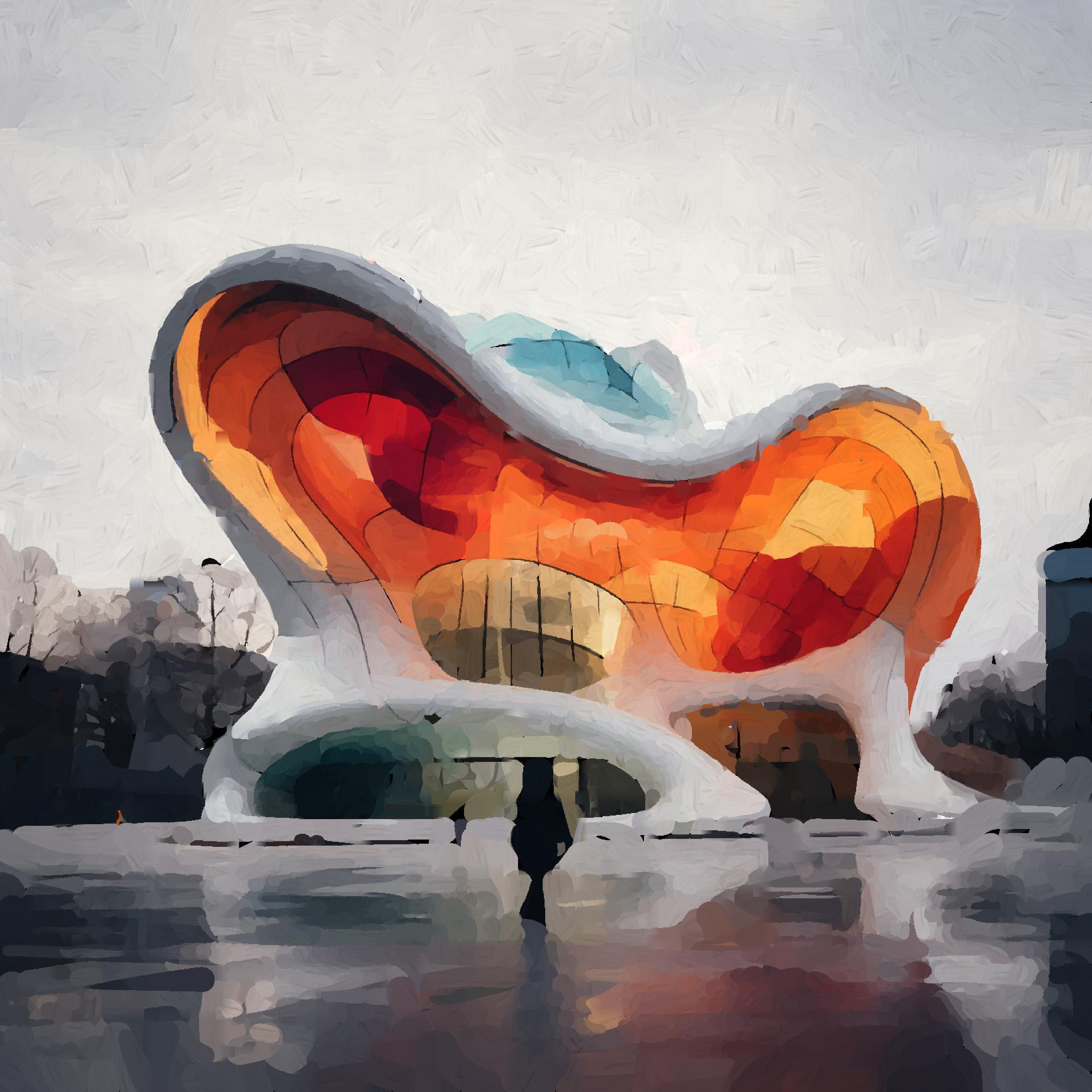}
\caption*{(c) Ours - Painterly}
\end{subfigure}


\caption{Painterly styles with a different degree of abstraction. This style generates organic human-like paintings, biasing the painting towards different structures.}
\label{fig:sota_comp_2}
\vspace{-1.5em}
\end{figure*}

%% file: figures/fig_tex/heatmaps.tex
\begin{figure}[ht]
\begin{center}
\includegraphics[width=\linewidth]{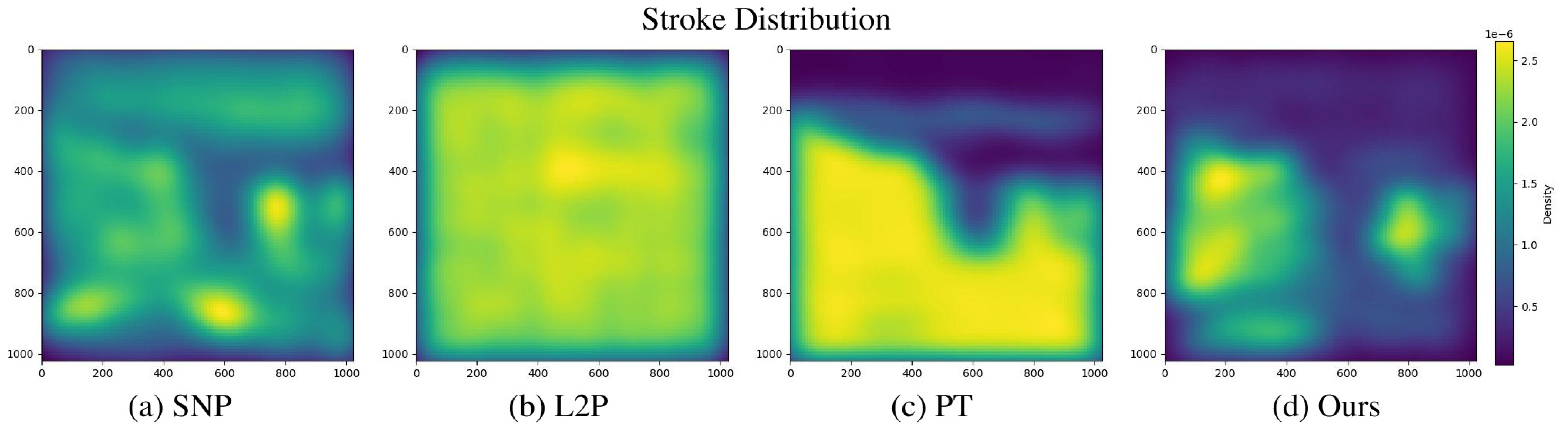}
\end{center}
\vspace{-1em}
\caption{Stroke distribution comparison of \Cref{fig:sota_comp_1} middle column. SNP (a) and PT (c) exhibit a disparse clustering, while L2P (b) paints evenly in a scene-agnostic way.
Our method (d) combines the selective focus of SNP with the clarity of PT, indicating a refined approach to applying strokes.}

\label{fig:heatmaps}
\vspace{-0.5em}
\end{figure}

%% file: tables/quant.tex
\begin{table}[b!]
  \centering
  \begin{adjustbox}{width=\linewidth, center}
  \begin{tabular}{l c c c c c c c c}
    \toprule
    \multirow{2}{*}{} & \multicolumn{2}{c}{Ours} & \multicolumn{2}{c}{SNP} & \multicolumn{2}{c}{PT} & \multicolumn{2}{c}{L2P} \\
    \cmidrule(r){2-3} 
    \cmidrule(r){4-5} 
    \cmidrule(r){6-7}
    \cmidrule(r){8-9}
      & $\mathcal{L}_1\downarrow$ & $\mathcal{L}_{perc} \uparrow$ & $\mathcal{L}_1\downarrow$ & $\mathcal{L}_{perc} \uparrow$        & $\mathcal{L}_1\downarrow$ & $\mathcal{L}_{perc} \uparrow$ & $\mathcal{L}_1\downarrow$ & $\mathcal{L}_{perc} \uparrow$\\
    \midrule
    Objects & \textbf{0.038} & \textbf{0.680} & 0.052 & 0.554 & 0.085 & 0.480 & 0.053 & 0.586 \\
    Buildings & \textbf{0.026} & \textbf{0.720} & 0.049 & 0.515 & 0.071 & 0.490 & 0.045 & 0.632 \\
    Streets & \textbf{0.031} & \textbf{0.709} & 0.041 & 0.582 & 0.072 & 0.486 & 0.041 & 0.620 \\
    Animals & \textbf{0.016} & \textbf{0.745} & 0.029 & 0.573 & 0.054 & 0.535 & 0.026 & 0.647 \\

    \bottomrule
  \end{tabular}
    \end{adjustbox}
    \vspace{0.2cm}
  \caption{Quantitative results on realistic style across different image domains. All methods use the same number of strokes (4000) on a 512x512 canvas.}
  \label{tab:quant}
  \vspace{-1em}
\end{table}

%% file: sec/5_conclusion.tex
\section{Conclusion}
\label{sec:conclusion}

This paper presents a semantic-based painting method that is more controllable, produces more visually appealing paintings, and achieves better realistic paintings than previous methods, as validated in the user studies and quantitative evaluations. Contrary to previous methods, our work offers a controllable framework that, besides realism, also aims for painterly and abstracted styles without the need for style-transfer like techniques. Furthermore, it is able to paint at any aspect ratio without further resizing steps. Future work includes incorporating a differentiable pasting module that enables general-canvas level loss, along with our patch loss strategy, for further exploration in stylization.



\input{tables/eff}

%% file: tables/eff.tex
\begin{table}[t]
\begin{adjustbox}{width=\linewidth, center}
  \centering
  \begin{tabular}{l c c c c}
    \toprule
    \multirow{2}{*}{Method} & \multicolumn{2}{c}{Optimization-based} &\multicolumn{2}{c}{Neural-based} \\
    \cmidrule(r){2-3} 
    \cmidrule(r){4-5} 
      & Ours & Optim \cite{Zou_2021_CVPR} & Transf \cite{Liu_2021_ICCV} & RL \cite{huang2019learning}\\
    \midrule
    
    \text{Res.} 1024 & \textbf{99.60s} & 447.10s & \textbf{0.514s} & 0.537s \\
    
    \midrule
    \text{Req. Training} & No & No & Yes & Yes \\
    
    \midrule
    \text{Use Dataset} & No & No & No & Yes  \\
    \bottomrule
  \end{tabular}
  \end{adjustbox}
  \vspace{0.2cm}
  \caption{Inference efficiency for different methods, measured in seconds for a single image at 1024x1024 and 1600 strokes.}
  \label{tab:eff}
\end{table}